\documentclass[preprint,12pt]{elsarticle}

\usepackage{geometry}
\usepackage{amssymb}
\usepackage{amsmath}
\usepackage{natbib}
\usepackage{xcolor}
\usepackage{hyperref}
\usepackage{graphicx}  
\usepackage{multirow}
\usepackage{algorithm}
\usepackage{algorithmic}
\usepackage{float}
\usepackage{enumitem}
\usepackage{makecell}
\usepackage{float}
\usepackage{booktabs}
\usepackage{placeins}

\journal{Computers and Electronics in Agriculture}

\begin{document}

\begin{frontmatter}



\title{Optimizing Navigation And Chemical Application in Precision Agriculture With Deep Reinforcement Learning And Conditional Action Tree}


\author[inst1]{Mahsa Khosravi} 
\author[inst2]{Zhanhong Jiang} 
\author[inst2]{Joshua R Waite} 
\author[inst3]{Sarah Jones}
\author[inst3]{Hernan Torres}
\author[inst3]{Arti Singh}
\author[inst2]{Baskar Ganapathysubramanian}
\author[inst3]{Asheesh Kumar Singh}
\author[inst2]{Soumik Sarkar} 
\affiliation[inst1]{organization={Department of Industrial and Manufacturing Systems Engineering},
            addressline={Iowa State University}, 
            city={Ames},
            state={Iowa},
            country={USA}}
\affiliation[inst2]{organization={Department of Mechanical Engineering},
            addressline={Iowa State University}, 
            city={Ames},
            state={Iowa},
            country={USA}}
\affiliation[inst3]{organization={Department of Agronomy},
            addressline={Iowa State University}, 
            city={Ames},
            state={Iowa},
            country={USA}}

\begin{abstract}
This paper presents a novel reinforcement learning (RL)-based planning scheme for optimized robotic management of biotic stresses in precision agriculture. The framework employs a hierarchical decision-making structure with conditional action masking, where high-level actions direct the robot's exploration, while low-level actions optimize its navigation and efficient chemical spraying in affected areas. The key objectives of optimization include improving the coverage of infected areas with limited battery power and reducing chemical usage, thus preventing unnecessary spraying of healthy areas of the field. Our numerical experimental results demonstrate that the proposed method, Hierarchical Action Masking Proximal Policy Optimization (HAM-PPO), significantly outperforms baseline practices, such as LawnMower navigation + indiscriminate spraying (Carpet Spray), in terms of yield recovery and resource efficiency. HAM-PPO consistently achieves higher yield recovery percentages and lower chemical costs across a range of infection scenarios. The framework also exhibits robustness to observation noise and generalizability under diverse environmental conditions, adapting to varying infection ranges and spatial distribution patterns.
\end{abstract}

\begin{graphicalabstract}
\includegraphics[width=1\textwidth]{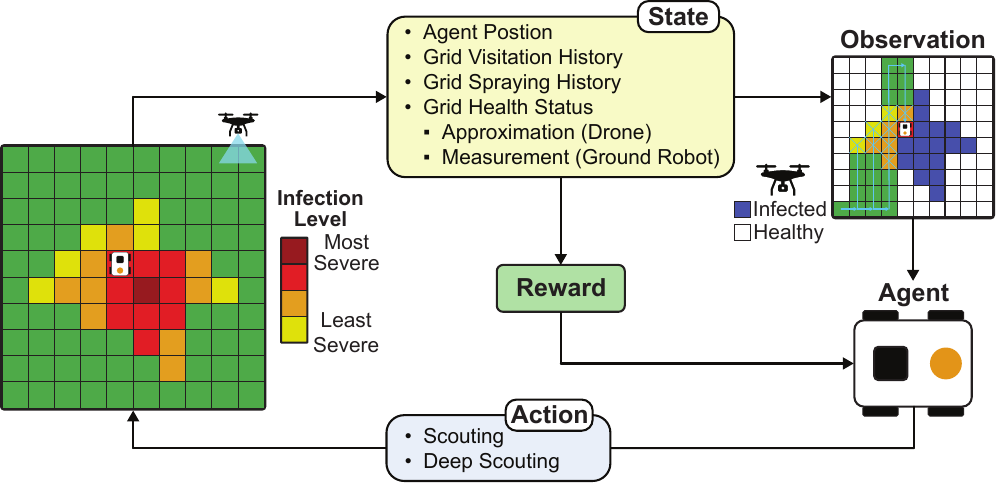}
\end{graphicalabstract}

\begin{highlights}
    \item We propose a novel RL framework—Hierarchical Action Masking Proximal Policy Optimization (HAM-PPO)—that integrates a hierarchical action space to coordinate efficient biotic stress scouting and targeted chemical spraying.
    \item We introduce a domain-specific reward mechanism that maximizes yield recovery while minimizing chemical usage by effectively handling noisy infection data and enforcing physical field constraints via action masking.
    \item We conduct a rigorous empirical evaluation across diverse, realistic biotic stress scenarios, capturing varying infection distributions and severity levels in row-crop fields.
    The proposed scheme is evaluated thoroughly, showing the framework’s effectiveness and robustness.
    \item Experimental results demonstrate that our approach significantly reduces non-target spraying, chemical consumption, and operational costs compared to baseline methods.

\end{highlights}


\end{frontmatter}



\section{Introduction}
\label{sec1}


The rapid advancement of technology has led to the emergence of precision agriculture, which addresses the growing global need for sustainable and efficient food production \cite{yaqot2023roadmap}. Precision agriculture is fundamentally about optimizing resource usage, improving crop health, and ensuring environmental sustainability \cite{rusmayadi2023revolutionizing}. A key component of achieving these goals is the precise application of agricultural inputs, particularly chemical (e.g., pesticides, herbicides and fungicides) across large and diverse agricultural sectors \cite{meshram2022pesticide}. Traditionally, schedule-based and uniform application approaches have resulted in excessive chemical use and negative environmental impacts. 
The emergence of Cyber-Agricultural Systems (CAS) that integrates AI-driven sensing, modeling, and actuation can alleviate such issues and enhance decision-making and resource efficiency in crop management \cite{sarkar2024cyber}. Leveraging these advancements, precision farming technologies, such as uncrewed aerial vehicles (UAVs) and ground robots, have emerged as powerful tools for targeted, site-specific pesticide spraying. Equipped with advanced navigation and spraying systems, these technologies enable efficient pesticide application, significantly reducing chemical inputs and minimizing environmental impact compared to baseline methods \cite{khan2021deep}. 

UAVs can be used for treating biotic stresses, offering advantages such as reducing farmers' exposure to harmful chemicals and covering large areas quickly \cite{kanase2018agriculture}. However, UAVs face challenges such as limited payload capacity and high maintenance costs due to the need to carry a chemical storage tank \cite{giles2015deployment}. In addition, challenges such as pesticide drift from high-altitude spraying still persist, leading to non-target contamination, environmental pollution \cite{martin2019effect}, and health risks for farm workers \cite{baharuddin2011pesticide, baldi2006pesticide, damalas2016farmers}. Furthermore, UAVs can sometimes struggle with accurate weed and insect detection, especially in complex environments \cite{giles2015deployment, martin2019effect}. Moreover, the widespread adoption of UAV-based spraying in agriculture is constrained by regulatory restrictions. For instance, limitations on beyond-visual-line-of-sight (BVLOS) operations pose significant barriers to large-scale agricultural spraying, reducing the feasibility of efficient coverage in extensive farming areas \cite{pathak2020use}. On the other hand, ground robots, such as AgBotII \cite{bawden2017robot} and BoniRob \cite{winterhalter2021localization}, have been developed to improve precision spraying in more localized environments. These systems combine LiDAR and camera-based sensors to perform localized spraying, improving maneuverability in challenging terrains. However, like UAVs, ground robots also encounter difficulties in accurately detecting weeds and insects, and they are affected by environmental factors such as weather and terrain conditions. Precise identification of insect species, encompassing both harmful pests and beneficial organisms, is crucial for implementing targeted pest control measures \cite{chiranjeevi2025insectnet}. As noted by \cite{malavazi2018lidar, higuti2019under}, ground robots, though highly maneuverable, may still experience challenges with accurate detection and adaptability in complex field conditions. A prominent example of recent advancements in precision pesticide spraying is John Deere’s See \& Spray technology \cite{avent2024comparing} which uses machine vision to detect and target weeds in real time. This technology has proven effective in reducing herbicide use while maintaining weed control. However, despite its success, See \& Spray still faces limitations related to environmental factors, weed size variability, and occasional inaccuracies in weed detection. False positives and false negatives in computer vision algorithms can result in over-spraying or under-spraying, thus reducing the system's efficiency in herbicide application. As the demand for reducing pesticide impact on the environment grows, more targeted and localized spraying methods are essential to accurately gather information and apply pesticides only where necessary, improving application precision.
However, a common challenge across these systems is the reliance on predefined decisions about spraying. In many of these systems, the decision to spray is typically made based on sensor data processed by computer vision algorithms, without considering economic or resource constraints, such as pesticide overuse and battery usage of electrified robots and drones.

Building on these advancements and motivated by the aforementioned gaps, our approach introduces an additional layer of decision-making that improves resource optimization in agricultural robots. Although existing technologies focus on optimizing pesticide application based on real-time detection, they still struggle with minimizing non-target spraying due to inaccuracies in pest or weed detection and lack of optimized planning. Leveraging reinforcement learning (RL), incorporates a hierarchical action space, allowing the agent to determine whether to perform a quick scouting survey or engage in more detailed deep scouting. Scouting, in this context, involves robots rapidly surveying the field 
to identify potential problem areas.
This step can provide a rough estimate of the problem areas in the field with minimal resource use. Meanwhile, deep scouting refers to a more thorough inspection of specific areas suspected of infection, during which the agent decides whether to spray based on the severity of the issue. By distinguishing between these two modes of operation, the hierarchical nature of our decision-making adds a further optimization level, enabling the agent to allocate resources more efficiently by improving time efficiency, saving battery power, and ensuring precise spraying—even in environments where computer vision might struggle with accuracy (during rapid scouting). This decision layer ensures that the robot only commits resources when necessary, whether for in-depth pest detection or treatment. Unlike existing technologies that depend on predefined routes and thresholds, our approach can operate in an unknown or partially-observed environment. For example, UAVs can be employed to conduct an initial scouting of the entire field, gathering information on potential areas of infection. 
Our method can integrate such an approximate infection map into the planning scheme, enabling the ground robot to navigate the field effectively. While we do not perform explicit path optimization, the system intelligently plans its movement to maximize both resource efficiency and navigation performance, ensuring that pesticide application is targeted and resource allocation is optimized throughout the process. Although this study focuses on optimizing pesticide application, the proposed framework can be extended to other precision chemical applications, such as targeted fertilization. Similar reinforcement learning-based strategies can be used to optimize nutrient distribution, enhancing resource efficiency and minimizing environmental impact.
Specifically, the main contributions of this work are outlined below.

\begin{itemize}[label=\textbullet]
    \item We develop a reinforcement learning (RL) framework, called  Hierarchical Action Masking Proximal Policy Optimization (HAM-PPO), that integrates a hierarchical action space to jointly optimize biotic stress scouting and targeted chemical spraying.
    
    \item We develop a domain-specific reward structure to maximize yield recovery while minimizing chemical usage. Our proposed RL framework effectively manages a noisy prior of the biotic stress distribution in field and utilizes action masking strategies to address the physical constraints inherent in row-crop agriculture. 
    
    \item We design a comprehensive empirical study that considers various realistic biotic stress distributions and severity levels in row-crop fields. 

    \item Using the empirical study, we demonstrate that our proposed framework can significantly reduce non-target spraying, chemical usage, and operational costs compared to various baseline approaches. 
    
    
\end{itemize}



\section{Literature review}
\label{Literature review}
\subsection{Precision spraying in agriculture}
Precision agriculture improves the efficiency of pesticide application through autonomous systems such as robots and UAVs,  as discussed in the reviews \cite{taseer2024advancements}, \cite{hanif2022independent}, \cite{meshram2022pesticide}, enabling targeted spraying to reduce chemical use and environmental impact. One of the integral aspects of achieving these benefits is the optimization of movement and spraying trajectories, ensuring that robots and drones navigate the most efficient paths for pesticide application. Several studies have focused on developing path-planning algorithms for pesticide spraying. To reduce operational cost, \cite{mahmud2019multi} presented the NSGA-III algorithm for multi-objective path planning in greenhouse pesticide spraying. The approach optimizes travel distance, turning angles, and pesticide capacity, modeling the task as a vehicle routing problem with the aid of a Probabilistic Roadmap (PRM) planner to determine optimal paths. This work was extended in~\cite{lal2017optimal} by addressing the optimization of robot paths under constraints such as pesticide carrying capacity and refilling at designated stations. This ensures that robots cover infected areas while maintaining appropriate pesticide levels for spraying, without minimizing the total pesticide used. Authors in~\cite{conesa2016route} proposed a route planning algorithm using Simulated Annealing, optimizing objectives such as distance, fuel, herbicide volume, input cost, and time, further improving operational efficiency in agricultural tasks like herbicide spraying.
In addition to navigation tasks, some researchers have focused on developing methods for reducing pesticide usage by designing variable rate spraying systems \cite{maghsoudi2015ultrasonic}, \cite{taseer2024advancements}. With advanced automation and robotics, this method can precisely reduce pesticide use, minimizing chemical exposure to farmers and lowering operational costs \cite{mahmud2020robotics}.
\subsection{RL for path planning and navigation in Ag robotics}
Efficient navigation in agricultural fields is inherently challenging due to the need to accurately estimate environmental uncertainties from limited sensor data \cite{gao2018novel}. Numerous studies have explored the use of RL for autonomous navigation and path planning in precision agriculture. For example, \cite{zhang2020whole} developed a whole-field RL method that allows UAVs to autonomously select flight paths, accurately predict crop health, and reduce energy consumption by scouting only a fraction of the field. Similarly, another study \cite{pourroostaei2021reinforcement} focused on optimizing UAV paths in smart farming through RL, specifically using Q-learning to autonomously explore the environment and identify cost-effective routes that minimize UAV power consumption and maximize data collection. Due to the differences and limitations between aerial and ground robots, many algorithms designed for UAVs are not directly applicable to ground-based vehicles. In this context, \cite{faryadi2021reinforcement} introduced a Dyna-Q+ based approach for unmanned ground vehicles (UGVs) to cooperatively model and cover unknown agricultural fields. Their method enables real-time learning, mapping plant rows, identifying obstacles, and defining regions of interest in dynamic farm environments. Authors in~\cite{weerakoon2024vapor} used offline RL to optimize legged robot navigation in unstructured outdoor environments, demonstrating the broader applicability of RL for navigating challenging terrains like dense vegetation. Further advancements in autonomous navigation for agricultural robots were made by \cite{yang2022intelligent} who introduced the Residual-like Soft Actor Critic (R-SAC) algorithm, enhancing dynamic obstacle avoidance and path planning for agricultural robots by integrating offline pre-training and a multi-step temporal difference mechanism, improving both path optimization and training efficiency. The challenge of unreliable GPS in agricultural environments has been addressed in~\cite{martini2022position}. They developed a position-agnostic navigation solution for vineyards using Deep Reinforcement Learning, allowing autonomous robots to navigate without precise localization to overcome GPS limitations in challenging environments. Additionally, for more specific task-oriented path planning in agricultural settings, \cite{huang2023automatic} explored the use of Deep RL (DRL) for optimizing pesticide spraying drones in complex orchard environments. Using Deep Q-Learning (DQN), the approach integrates environmental factors like tree density and pest infestation to automatically select efficient spraying paths, reducing pesticide use and operational costs.

\subsection{RL for resource optimization in precision agriculture}
RL has gained attention for optimizing decision-making in precision agriculture, particularly in resource management tasks like weeding, crop management, and irrigation, aiming to improve efficiency and reduce resource use. For instance, \cite{mcallister2021agbots} applied RL to optimize mechanical weeding using a bandit-based approach that coordinates robots to predict weed growth and allocate resources efficiently. This study highlights the integration of decision-making, resource allocation, and task-specific planning in agriculture. Similarly, \cite{gautron2022reinforcement} explored the role of RL in managing agricultural operations such as pest control and fertilization. Their review highlights RL's potential to adapt dynamically to fluctuating conditions, offering significant benefits for precision agriculture decision support systems (DSS). In the context of irrigation, \cite{chen2021reinforcement} demonstrated the application of DQN to optimize irrigation practices for paddy rice crops. By using weather forecasts to make real-time irrigation decisions, their approach significantly reduced water usage while maintaining crop yields. 
\subsection{Current research gaps}
The current literature on path planning for pesticide application mainly focuses on UAVs, with limited research on ground robots. Additionally, most studies address navigation and path optimization separately from pesticide usage minimization. No existing work has simultaneously optimized navigation for efficient coverage while minimizing chemical usage, nor has it integrated a third layer of optimization: determining when and where to apply pesticide. To address this gap, we propose a planning framework using the Proximal Policy Optimization (PPO) algorithm and a Conditional Action Tree structure, which facilitates efficient policy model outputs. This approach reduces the action space complexity through action parameterization and invalid action masking, enhancing both navigation and decision-making processes for pesticide application, especially in partially known crop rows.
\section{Preliminaries and problem formulation}
In this section, we present a brief technical background that characterizes the subsequent problem formulation and our proposed scheme. Deploying a mobile robot in a crop field to intelligently spray chemicals is an extremely challenging task, which comprises two difficulties. The first is to decide when to strategically deploy the robot in the field after the infection appears, such that the yield recovery can be maximized. The second challenge is to optimally navigate the robot to infectious areas to ensure that the robot can cover all the infectious parts of the field within its battery life. In addition to navigation, the robot also has to make the decision to spray chemicals, without incurring excessive pesticide costs. Separately solving these issues to get the optimal decision could be practically infeasible as optimizing one objective does not yield the optimal solution for the other ones. Moreover, the decision-making process needs to be \textit{dynamic} to allow the robot to take optimal actions quickly given the present state. Such decision-making problems are usually formulated as a Markov Decision Process (MDP) that can be solved using well-established algorithms. However, fully observable states of environment and/or robot are \textit{rare} in real-world problems. For example, if the robot obtains an accurate crop health status map of the entire field, it could significantly help with navigation. Unfortunately, before the robot reaches a specific part of the field with biotic stress, it is difficult to obtain precise information about the stress, such as the type, extent and severity of the problem. Even if the stress state can be estimated in some other way, such as from images taken by a drone, the state would still be imperfect and incomplete, that is, partially observable. Therefore, we formulate the problem as a Partially Observable MDP (POMDP), which is presented in the sequel.

\subsection{Partially Observable Markov Decision Process}
Formally, the POMDP model is defined as a 7-tuple $\mathcal{M}=\langle\mathcal{S},\mathcal{A}, \mathcal{T}, \mathcal{R}, \Omega, \mathcal{O}, \gamma\rangle$, where the variables are defined as follows:
\begin{itemize}
    \item $\mathcal{S}$ denotes the state space -- the set of all possible states, which can include the states of the robot and/or environment.
    \item $\mathcal{A}$ denotes the action space -- the set of all actions the robot can perform.
    \item $\mathcal{T}(s,a,s')$ denotes the transition function, representing the nondeterministic effects of the actions. It is a conditional probability function such that $\mathcal{T}(s,a,s')=\mathbb{P}[S_{t+1}=s'|S_t=s, A_t=a]$, representing the probability that the robot is at state $s'\in\mathcal{S}$ after executing action $a\in\mathcal{A}$ at state $s\in\mathcal{S}$. $t$ denotes an arbitrary time step.
    \item $\mathcal{R}$ denotes the immediate reward function. This function can be parameterized by a state, a state–action pair, or a tuple of state, action, and subsequent state. In this context, it is defined as $\mathcal{R}(s,a)=\mathbb{E}[R_{t+1}|S_t=s,A_t=a]$.
    \item $\Omega$ denotes the observation space -- the set of all observations the robot receive.
    \item $\mathcal{O}(s',a,o)$ denotes the observation function, indicating errors and noise in measurement and perception. It is also a conditional probability $\mathcal{O}(s',a,o)=\mathbb{P}[O_{t+1}|S_{t+1}=s', A_t=a]$, indicating the observation the robot may receive when it is at the state $s'\in\mathcal{S}$ after executing the action $a\in\mathcal{A}$. 
    \item $\gamma$ denotes the discount factor between $[0,1]$.
\end{itemize}
Specifically, a POMDP $\mathcal{M}$ will operate as follows. At an arbitrary time step $t$, the agent is at some state $S_t=s\in\mathcal{S}$. Different from MDP, where $s$ is fully observable, in this context, $s$ is partially observable, which can be attributed to inaccessible measurements or only observations of part of the state space. 
Given the current state $s$, the agent infers the best action $A_t=a\in\mathcal{A}$ to execute such that $S_t$ transits to a new state $S_{t+1}=s'\in\mathcal{S}$ based on the transition dynamics $\mathcal{T}$. Essentially, $s'$ is hidden from the agent, while the agent perceives an observation $O_{t+1}=o\in\Omega$ that discloses some information about $s'$ respecting the observation function $\mathcal{O}$. Finally, a reward $R_{t+1}=r$ is received by the agent via the reward function $\mathcal{R}$. Typically, both $\mathcal{T}$ and $\mathcal{R}$ are unknown to the agent, and in this work, we only specify $\mathcal{R}$ based on the domain knowledge. Deep RL algorithms aim at searching for an optimal policy that maximizes the cumulative discounted reward, which can be denoted by $G_t=\sum_{k=0}^\infty\gamma^kR_{t+k+1}$. Compared to MDP that has been well studied and established in a principled manner~\cite{puterman1990markov}, POMDP is fairly underexplored, such that most proposed methods for MDP may not be directly applicable to problems formulated as POMDP. Developing approaches to solving POMDP effectively still remains extremely challenging. Therefore, in order to leverage existing efficient algorithms, POMDP will degenerate to MDP by resorting to observations to estimate the hidden state, which will be presented in the problem formulation. To conclude this subsection, we introduce the key value functions, i.e., state value and state-action value functions. Formally, we denote by $\pi:\mathcal{S}\to\mathcal{A}$ the policy function. Thus, the state value function is defined as $V^\pi(s)=\mathbb{E}[G_t|S_t=s]$, which is the expected return starting from an arbitrary state. Similarly, the state-action value function is defined as $Q^\pi(s,a)=\mathbb{E}[G_t|S_t=s,A_t=a]$, indicating the expected return of taking an action starting from an arbitrary state. The advantage function $\mathbf{A}$ is defined as $\mathbf{A}^\pi(s,a)=Q^\pi(s,a)-V^\pi(s)$, which informs how good it is by taking the action $a$ compared to the average. 
\subsection{Problem formulation}
This work focuses primarily on developing optimal decision-making strategy for an agricultural robot to apply pesticides in a field for recovering yield, while minimizing pesticide costs. Concretely, we consider a field with healthy and infectious crops, which can be divided into multiple grids. To simplify the analysis, we assume that each grid is either healthy or infectious. In practice, inside each grid, there may exist some infectious/healthy crops simultaneously. To formulate this problem into a POMDP, we now define the state, action, and reward in the following. 
\subsubsection{State space}
The state space in an RL problem can include the state of the environment and/or the agent. Unlike many existing approaches that focus either solely on optimal path planning for coverage tasks or on pesticide application without efficient navigation, our approach addresses both challenges by incorporating information from both the robot and the environment in the state space. Specifically, $\mathcal{S}$ is defined as follows:
\begin{itemize}
    \item The health status of the field, denoted by $s_1$, is modeled as a grid matrix where each grid indicates whether it is healthy or infected. Each infected grid is assigned one of three discrete infection levels, denoted as \( I_1 \) (mild), \( I_2 \) (moderate), and \( I_3 \) (severe), corresponding to the severity of pest infestation. Healthy grids are denoted as \( h \), indicating they are free from infection. The field is structured into grids with dimensions \( l \times w \), where \( l \) and \( w \) represent the length and width of the field, respectively. 
    The health status of any grid \( (i, j) \) is represented by \( H_{i,j} \).
    Thus, the health status of the entire field can be represented as a matrix:

    \begin{align}
        H_{l \times w} &=
        \begin{bmatrix}
        H_{1,1} & \cdots & H_{l,1} \\
        \vdots & \ddots & \vdots \\
        H_{1,w} & \cdots & H_{l,w}\\
        \end{bmatrix},
    \end{align}
    \\
    The fields outside the grid, where the robot has not yet entered, are labeled as \( e \), representing ``empty" areas (free of vegetation) where the agent can start and commute. In this work, the robot begins at the field’s edges, where grids are labeled as ``empty". This also complies with the real-world deployment of ground robots.
    \item The agent’s position within the field, \( (i, j) \), denoted by $s_2$, indicates the current location of the robot within the field.
    \item The history of visitation, denoted by $s_3$, records which grids have been visited by the robot.
    \item The history of pesticide spraying, denoted by $s_4$, tracks the spraying actions performed by the robot on each grid, enabling the robot to avoid redundant actions and optimize its operations.
\end{itemize}
\subsubsection{Action space}
We formulated the action space using a parameterized structure, as introduced in \cite{masson2016reinforcement} for the RoboCup 2D Half-Field-Offense environment. This structure, extended into a hierarchical framework for precision agriculture, is defined as a \textit{Hierarchical Action Space} and consists of an action type and a discrete parameter, described by action trees with two components: \(a = \{b_0, b_1\}\). Here, \(b_0\) represents the high-level action type (scouting or deep scouting), and \(b_1\) is the discrete low-level action parameter specifies the sub-action based on the chosen high-level type. If $b_0$ is scouting, then $b_1$ indicates movement in one of four directions (\textit{up}, \textit{down}, \textit{left}, or \textit{right}). If $b_0$ is deep scouting, $b_1$ dictates whether or not to spray. 
The possible actions are moving in any direction, deep scouting without spraying, and deep scouting with spraying. 

In precision agriculture, scouting and deep scouting are high-level tasks that allow the agent to assess crop health at different levels of detail.
This hierarchical action structure allows the agent to choose between scouting or deep scouting first, and then perform the corresponding low-level action (movement or spraying) based on its assessment.
\subsubsection{Reward}
The reward structure for the robot’s actions is designed to reflect the potential yield recovery and penalties based on the actions performed during scouting and spraying. The concepts of attainable yield and yield loss, as defined in \cite{khosravi2024aggym}, are taken from our previous work, where we explored the effects of infection on crop yield. These two concepts are the critical parameters in the reward structure induced by scouting and deep scouting. Potential yield represents the maximum crop production achievable under ideal conditions, influenced by factors such as sunlight, temperature, and rainfall. Attainable yield accounts for limitations like water and nutrient availability, while yield loss incorporates reductions caused by pathogens, pests, and weeds. The infection duration and severity are key parameters influencing yield loss. This foundational understanding forms the basis for defining the reward structure.

\noindent\textbf{Attainable Yield fraction.} The attainable yield fraction is represented by the following formula:
\begin{equation}
\eta_Y = \frac{e^{-(T_{\text{inf}} + T_{50\%})}}{1 + e^{-(T_{\text{inf}} + T_{50\%})}}
\label{eq:attainable_yield}
\end{equation}
where $T_{50\%}$ signifies the infection duration by which there will be $50\%$ of the total yield loss and ($T_{inf}$)
represents the infection duration, which is the number of days since the initial infection. This duration is directly related to the severity of the pest infestation and helps quantify how the length of infection influences the potential yield loss.

\noindent\textbf{Yield Loss.}
The yield loss is defined as:
\begin{equation}
\text{Yield Loss} = (1 - \eta_Y) \times p_{\text{inf}} \label{eq:yield_loss}
\end{equation}
where \( p_{\text{inf}} \) is the normalized severity index, with values in the range \([0, 1]\) representing the infection's severity level. We now proceed to introduce the individual components of the reward function.

\noindent\textbf{Scouting.} To ensure the efficient navigation, we consider the reward components induced by the scouting. The first one is
\textit{Visiting Infected Unit}.
The reward for visiting an infected unit is based on the potential yield recovery if the infection is addressed immediately:
\begin{equation}
r_{\text{sc\_inf}} = \eta_Y \times UAY \times PPB \times PS
\end{equation}
where \( UAY \) is the unit attainable yield, \( PPB \) is the price per bushel of yield, and \( PS \) is the probability of spraying that unit in the following action (the likelihood of choosing spraying as the action parameter after visiting the infected unit). The second one is
\textit{Revisiting Healthy Unit}.
A penalty of $-\kappa_{rev}$ is applied when revisiting a healthy unit, where $\kappa_{rev}$ represents a predefined penalty parameter for redundant visits. 
\[
r_{sc\_healthy} = -\kappa_{rev}
\]

\noindent\textbf{Deep Scouting.} In addition to scouting, deep scouting is taken into consideration in the reward function, which assists in enabling more efficient pesticide spraying. The first key component in the deep scouting is
\textit{Spraying Infected Unit}.
If the agent chooses to spray an infected unit, the reward is calculated as:
\begin{equation}
r_{\text{dsc\_spray\_inf}} = \eta_Y \times UAY \times PPB - UPP
\end{equation}
where \( UPP \) is the unit pesticide price.
Another key component is
\textit{Not Spraying Infected Unit}.
If the agent chooses not to spray an infected unit, it incurs a penalty based on the yield loss it could have prevented by spraying:
\begin{equation}
r_{\text{dsc\_nospray\_inf}} = - \text{Yield Loss} \times UAY \times PPB
\end{equation}
The last key component is
\textit{Spraying Healthy Plot}.
A penalty equivalent to the cost of the pesticide is applied:
\begin{equation}
r_{\text{dsc\_spray\_healthy}} = - UPP
\end{equation}
\noindent\textbf{Total Reward.}
The total reward \( r_{\text{total}} \) for the agent is the sum of the individual rewards from scouting and deep scouting actions:
\begin{equation}
r_{\text{total}} = r_{\text{sc\_inf}} + r_{\text{sc\_healthy}} + r_{\text{dsc\_spray\_inf}} + r_{\text{dsc\_nospray\_inf}} + r_{\text{dsc\_spray\_healthy}}
\end{equation}
\subsection{Termination Condition}
In our formulation, the (battery) power constraint of the robot (\(E_t\)) serves as a termination condition that limits the agent’s actions during the episode. Power consumption depends on the action type:
{\begin{itemize}
    \item \textbf{Movement (Scouting)}: Requires relatively low power and completes in a duration of \( \tau_s \).
    \item \textbf{Spraying (Deep Scouting)}:  Requires significantly more power and takes a longer duration of \( \tau_d \), as it involves detailed stress assessment before making a spraying decision.
\end{itemize}}
The episode terminates when the power source (e.g., battery) is depleted, ensuring that the agent must manage its power source usage efficiently to balance task performance with resource constraints.

\subsection{From POMDP to approximate MDP} In Section 3.1, we have introduced POMDP to characterize our problem formulation. Nevertheless, solving POMDP efficiently still remains a challenging task, which makes it underexplored without numerous well-established methods. In contrast, MDP has been well studied with many mature approaches, such as value-based and policy-based algorithms. Previous work~\cite{jiang2021building} has proposed to degenerate POMDP to MDP through state estimation, which can be done by leveraging state observations. With this, popular and powerful RL algorithms can be used to solve the approximate MDP problem in this context. 
In the sequel, we present the technical details of how the POMDP degenerates to the MDP.

\begin{figure}
    \centering
    \includegraphics[width=0.5\linewidth]{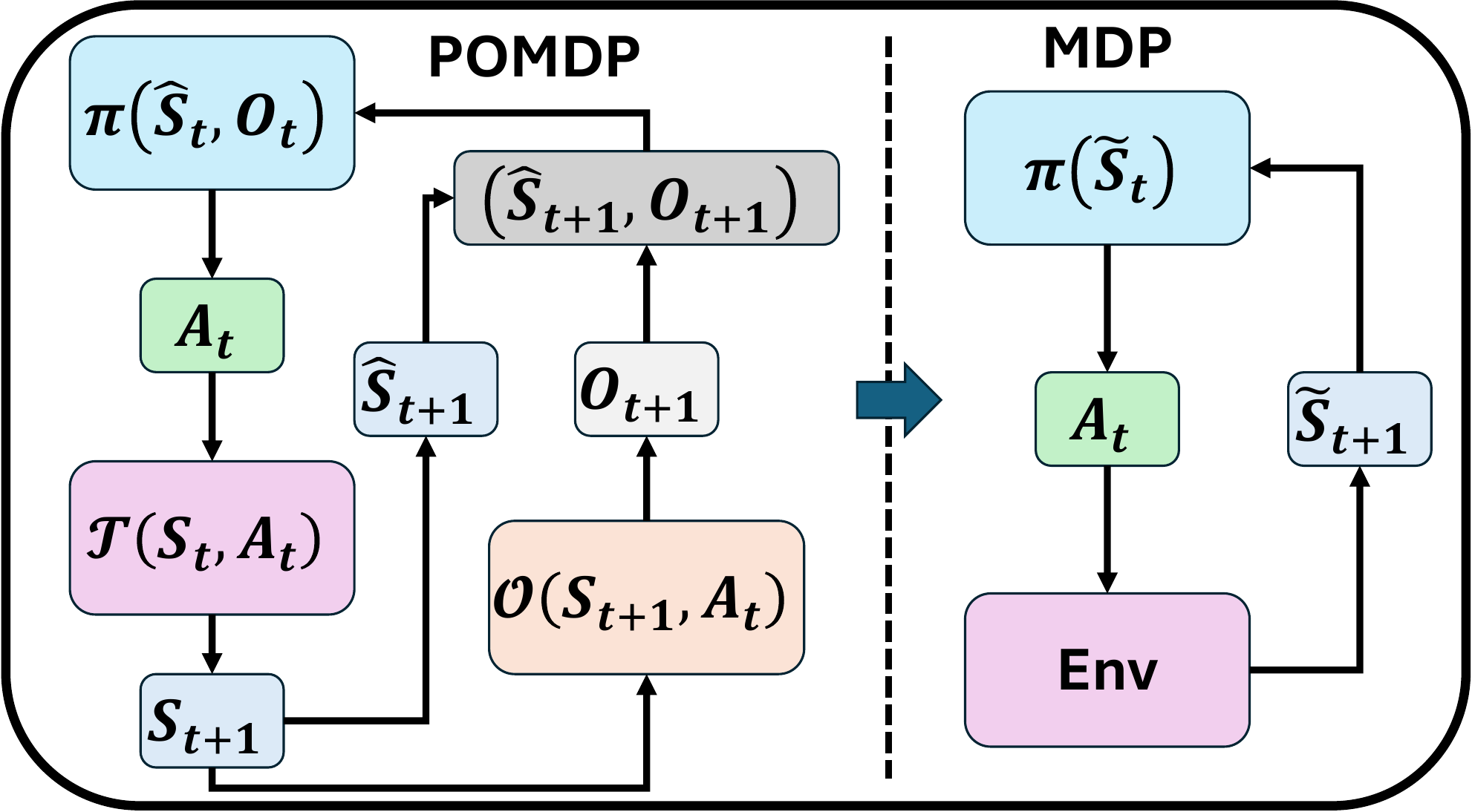}
    \caption{Degeneration from POMDP to MDP through state estimation. For the purpose of illustration, we skip the reward from the environment to the agent.}
    \label{fig:pomdp-mdp}
\end{figure}

As described earlier, the state space $\mathcal{S}$ is defined as $\mathcal{S}=\{s_1, s_2, s_3, s_4\}$. $s_2$ indicates the robot's position in the field, which is fully observable since the field map with respect to the crop planting distribution remains fixed during the robot's operation. Hence, we assume that the position of the robot within the field at any time step can be determined using accurate GPS measurements. Similarly, the history of visitation $s_3$ and the history of pesticide spraying $s_4$, are also precisely known by the robot. At an arbitrary step, when the robot reaches a new grid, it can immediately get access to the specific infection level of that grid. However, the robot has no knowledge about the precise infection levels of those grids the robot has not visited. Hence, the robot will always have the \textit{partial} knowledge of $s_1$ before covering all grids in the field, which justifies the introduction of POMDP. In practice, before deploying autonomous robots to spray pesticides, preliminary information about infections in the field can be roughly assessed using images captured by drones. These observations can help guide the robot’s navigation and determine the appropriate amount of pesticide to spray. While precise knowledge of the infection across the entire field may not be available, prior observations provide an estimate that aids in decision-making. Notably, if the quality of these observations improves significantly for instance, if UAV-based scouting methods become highly accurate and provide a complete and noise-free field assessment then the problem would become fully observable. In that case, our POMDP formulation would naturally degenerate into an MDP, since all state variables would be directly accessible to the agent. However, our framework remains valid under both conditions: with partial observability, it exploits hierarchical decision-making to refine uncertainty and improve decision-making; and with full observability, the same hierarchical structure can still be leveraged to efficiently direct the robot toward infected regions, now relying on complete information rather than uncertainty reduction. Also note that in this study, the evolution of infection within the field occurs in a slow time scale. Therefore, the infection distribution is assumed to be \textit{static} during the robot's operation. However, the accuracy of the observation improves since the health status of each grid is revealed when the robot visits that grid.
Figure~\ref{fig:pomdp-mdp} shows how the POMDP formulation degenerates to an MDP in our study through the state observation. 
\FloatBarrier

We denote the true and estimated states by $S_t$ and $\tilde{S}_t$ respectively at the time step $t$. Based on the above discussion, $\tilde{S}_t$ can be decomposed as the exact state information of $s_2, s_3, s_4$ and the partial state information of $s_1$. For the convenience of analysis, the full state information of $s_2, s_3, s_4$ at $t$ is represented by $\hat{S}_t$, while the observation of $s_1$ at time step $t$ is $O_t$. Immediately, we have $\tilde{S}_t=\hat{S}_t\bigwedge O_t$. As shown in Figure~\ref{fig:pomdp-mdp}, given the current state $S_t$, which is unknown to the agent, the agent executes an action $A_t$ that leads to the state transition from $S_t$ to $S_{t+1}$ through the underlying transition dynamics $\mathcal{T}(S_t, A_t)$. Subsequently, the state estimator $\mathcal{O}(S_{t+1}, A_t)$ produces an observation $O_{t+1}$, particularly for the health status. As we are still able to access $\hat{S}_{t+1}$, the policy $\pi$ takes as input $\tilde{S}_{t+1}= \hat{S}_{t+1} \bigwedge O_{t+1}$. In this context, both $\mathcal{T}$ and $\mathcal{O}$ are unknown to the agent, which can be regarded as one entire component inside the environment. Hence, the POMDP is now approximated as an MDP that can be solved by many well-known RL algorithms, one of which is introduced in the following.
\subsection{Proximal Policy Optimization}
In our study, we resort to proximal policy optimization (PPO) algorithm for training the RL agent to maximize the cumulative discounted rewards. In what follows, we review the algorithm for characterizing the developed framework. 

PPO has become a default baseline in a variety of applications~\cite{dai2023augmented,alagha2022target,wu2024combustion}. It is favored due to its exceptional performance and simple implementation with theoretically sound foundation given by the policy improvement lower bound. Intuitively speaking, PPO seeks to constrain the updated policy close to the present one with a \textit{clipping} heuristic, which results in a widespread variant called PPO-clip~\cite{huang2024ppo}. Particularly, the following objective is solved at every policy update:
\begin{equation}\label{eq_1}
    \mathcal{L}^{clip}_t(\pi)=\mathbb{E}_{(s,a)\sim d^{\pi_t}}[\textnormal{min}(\frac{\pi(a|s)}{\pi_t(a|s)}\hat{\mathbf{A}}^{\pi_t}(s,a),\textnormal{clip}(\frac{\pi(a|s)}{\pi_t(a|s)},1-\epsilon,1+\epsilon)\hat{\mathbf{A}}^{\pi_t}(s,a))],
\end{equation}
where $\textnormal{clip}(a,b,c)=\textnormal{min}(\textnormal{max}(a,b),c)$. The clipping function plays a central role in this objective as it consistently enforces the probability ratio between the current and next policies in a reasonable range between $[1-\epsilon, 1+\epsilon]$, where $\epsilon$ is clipping hyperparameter. $\frac{\pi(a|s)}{\pi_t(a|s)}$ is larger than 1 if $a$ is more likely to be taken in the current state $s$ after the policy update, and smaller than 1 if the opposite is true. $\hat{\mathbf{A}}^{\pi_t}$ represents an estimator of the advantage function $\mathbf{A}^{\pi_t}$. In this paper, we use Generalized Advantage Estimate (GAE)~\cite{schulman2015high} to estimate $\hat{\mathbf{A}}^{\pi_t}$. For each update, PPO use a fixed length horizon trajectory $\mathcal{H}$, which runs the policy for $\mathcal{H}$ time steps to collect experiences. Therefore, the estimator $\hat{\mathbf{A}}^{\pi_t}$ is given by:
\begin{equation}\label{eq_2}
    \hat{\mathbf{A}}^{GAE(\gamma,\lambda)}_t=\sum^\mathcal{H}_{l=0}(\gamma\lambda)^l\delta_{t+l},
\end{equation}
where $\delta_{t+l}=r_{t+l}+\gamma V^{\pi_t}(s_{t+l+1})-V^{\pi_t}(s_{t+l})$ is the temporal difference residual, $\gamma, \lambda\in[0,1]$ are discount factors, and $V^{\pi_t}(s_t)$ is the value typically predicted by the value network (critic). One intuitive implication from Eq.~\ref{eq_1} is that the term $\frac{\pi(a|s)}{\pi_t(a|s)}\hat{\mathbf{A}}^{\pi_t}(s,a)$ motivates the policy to select actions that yield high positive advantages, while the term $\textnormal{clip}(\frac{\pi(a|s)}{\pi_t(a|s)},1-\epsilon,1+\epsilon)\hat{\mathbf{A}}^{\pi_t}(s,a)$ sets a constraint for $\frac{\pi(a|s)}{\pi_t(a|s)}$ to remove the incentive of moving outside the interval $[1-\epsilon, 1+\epsilon]$, which ensures the updated policy to not deviate far from the current policy and the learning stability. 

Apart from the policy loss, in PPO, additional loss terms consisting of value function loss $\mathcal{L}^V_t(\pi)$ and the entropy loss with respect to the policy $\pi(a|s)$, $E_t(\pi)$, are combined with Eq.~\ref{eq_1} such that entire objective loss is expressed as follows:
\begin{equation}
    \mathcal{L}_t(\pi)=\mathcal{L}^{clip}_t(\pi)-c_1\mathcal{L}^V_t(\pi)+c_2E_t(\pi).
\end{equation}
$c_1$ and $c_2$ are the tunable coefficients. 
$\mathcal{L}^V_t(\pi)=(V^{\pi}(s_t)-V^{target}(s_t))^2$, where $V^{target}(s_t)$ is the target state value. $E_t(\pi)$ ``encourages" the policy exploration. Though PPO is well-known for its balance between sample efficiency and wall-clock time, particularly when compared to other policy optimization algorithms such as TRPO~\cite{schulman2015trust}, in practice, the actions executed may be subject to constraints or restrictions. For example, an agricultural robot in the field may not be allowed to move randomly and needs to follow some rules, which result in invalid actions in $\mathcal{A}$. This motivates us to study a formal mechanism called \textit{invalid action masking} that is introduced in the following. 

\subsection{Invalid Action Masking (IAM)}
Invalid action masking has remained a widespread technique implemented to avoid sampling invalid actions in large action space~\cite{huang2020closer,ye2020mastering}. Though in most existing RL works, IAM has not received considerable attention, it plays a critical role in invalidating some actions in the action space $\mathcal{A}$. To illustrate how IAM works, we use a toy example and simplify the action space $\mathcal{A}$ to a discrete one with only five actions under two states. We employ a neural network to represent the policy in PPO, which outputs unnormalized score (logits) and then converts them into an action probability distribution via a softmax operation. Let $\mathcal{A}=\{a_0,a_1,a_2,a_3,a_4\}$ and $\mathcal{S}=\{s_0,s_1\}$. Suppose that the environment state transits to the terminal state after one action is executed in the initial state $s_0$ and that the RL agent receives an immediate reward $r=1$. Consider a policy $\pi_\theta$ that is parameterized by $\theta=[\theta_0,\theta_1,\theta_2,\theta_3,\theta_4]=[1.0,1.0,1.0,1.0,1.0]$. To simplify the analysis, we also assume that the policy directly outputs $\theta$ as the output logits. Thus, we have:
\begin{equation}
\begin{split}
    \pi_\theta(\cdot|s_0)&=[\pi_\theta(a_0|s_0),\pi_\theta(a_1|s_0),\pi_\theta(a_2|s_0),\pi_\theta(a_3|s_0),\pi_\theta(a_4|s_0)]\\&=\text{softmax}([\theta_0,\theta_1,\theta_2,\theta_3,\theta_4])\\&=[0.2,0.2,0.2,0.2,0.2].
\end{split}
\end{equation}
In this context, $\pi_\theta(a_i|s_0)=\frac{\text{exp}(\theta_i)}{\sum_j\text{exp}(\theta_j)}$. Assume that action $a_0$ is sampled from $\pi_\theta(\cdot|s_0)$ such that policy gradient~\cite{nota2019policy,sutton1999policy} is calculated as follows:
\begin{equation}
\begin{split}
    g&=\mathbb{E}_\tau[\nabla_\theta\sum_{t=0}^T\text{log}\pi_\theta(a_t|s_t)G_t]\\&=\nabla_\theta\text{log}(a_0|s_0)G_0\\&=[0.8,-0.2,-0.2,-0.2,-0.2],
\end{split}
\end{equation}
where 
\[
(\nabla_\theta\text{log softmax}(\theta)_j)_i=\begin{cases}
    1-\frac{\text{exp}(\theta_i)}{\sum_j\text{exp}(\theta_j)} & \text{if $i=j$} \\
    -\frac{\text{exp}(\theta_i)}{\sum_j\text{exp}(\theta_j)} & \text{Otherwise}
\end{cases}
\]
In this context, suppose that $a_1$ is invalid for state $s_0$ such that the valid actions are $a_0,a_2,a_3,a_4$. The IAM assists in sampling invalid action from the $\pi_\theta$ by ``masking out" those logits in the output corresponding to the invalid actions. The masking can be realized mathematically by replacing the logits of invalid actions with large negative numbers $C$ (e.g., $C=-1\times 10^{10}$). Denote by $h(\cdot):\mathbb{R}\to\mathbb{R}$ the \textit{masking} function. Hence, the action probability distribution has now become
\begin{equation}
\begin{split}
    \pi'_\theta(\cdot|s_0)&=\text{softmax}(h([\theta_0,\theta_1,\theta_2,\theta_3,\theta_4]))\\&=\text{softmax}([\theta_0,C,\theta_2,\theta_3,\theta_4])\\&=[\pi'_\theta(a_0|s_0),\varepsilon,\pi'_\theta(a_2|s_0),\pi'_\theta(a_3|s_0),\pi'_\theta(a_4|s_0)]\\&=[0.25,0.,0.25,0.25,0.25],
\end{split}
\end{equation}
where $\varepsilon$ is small probability of the masked invalid action. When $C$ increases, $\varepsilon$ approaches 0. Denote by $g'$ the policy gradient with IAM such that $g'=[0.75,0,0.25,0.25,0.25]$, which suggests that IAM ensures the policy gradient value corresponding to the logits of the invalid action to be zero. However, one may argue that if $g'$ could be a policy update for the masked policy $\pi'_\theta$. An existing work~\cite{huang2020closer} has provably shown this as a formal result, which also sets the theoretical foundation in our work.

\subsection{Conditional Action Tree (CAT)}
Generically speaking, each action in $\mathcal{A}$ can possibly be sampled and executed with equal significance when RL agent explores. Nevertheless, in some scenarios, based on the given state, the agent selects an action that inherently minimizes the policy model's output size. This approach simultaneously enables action parameterization and simplifies action reduction through the IAM mechanism described earlier. Intuitively, the action space is framed as a action hierarchy that helps in establishing an action tree. For example, when an agricultural robot is deployed to apply pesticides in a field, the required actions qualitatively comprise navigation and spraying pesticides. In this study, the navigation is simplified to have a few \textit{discrete} actions, i.e., \textit{up, down, left, right}. Spraying pesticides also only involves two actions, i.e., \textit{spray, not spray}, which resembles a binary-class action. One can define $\mathcal{A}$ to include all these actions and learn the policy as usual. However, there exists an implicit hierarchy in the action space such that we resort to an action tree instead for sampling and executing action during policy learning. Additionally, some actions may not be valid or cannot be performed given a state. Thereby, an action tree can decompose the whole action space to ensure that the IAM is implemented in a sub-tree, instead of the whole tree. In what follows, we mathematically define the conditional action tree based on~\cite{bamford2021generalising}.
\begin{figure}
    \centering
    \includegraphics[width=0.5\linewidth]{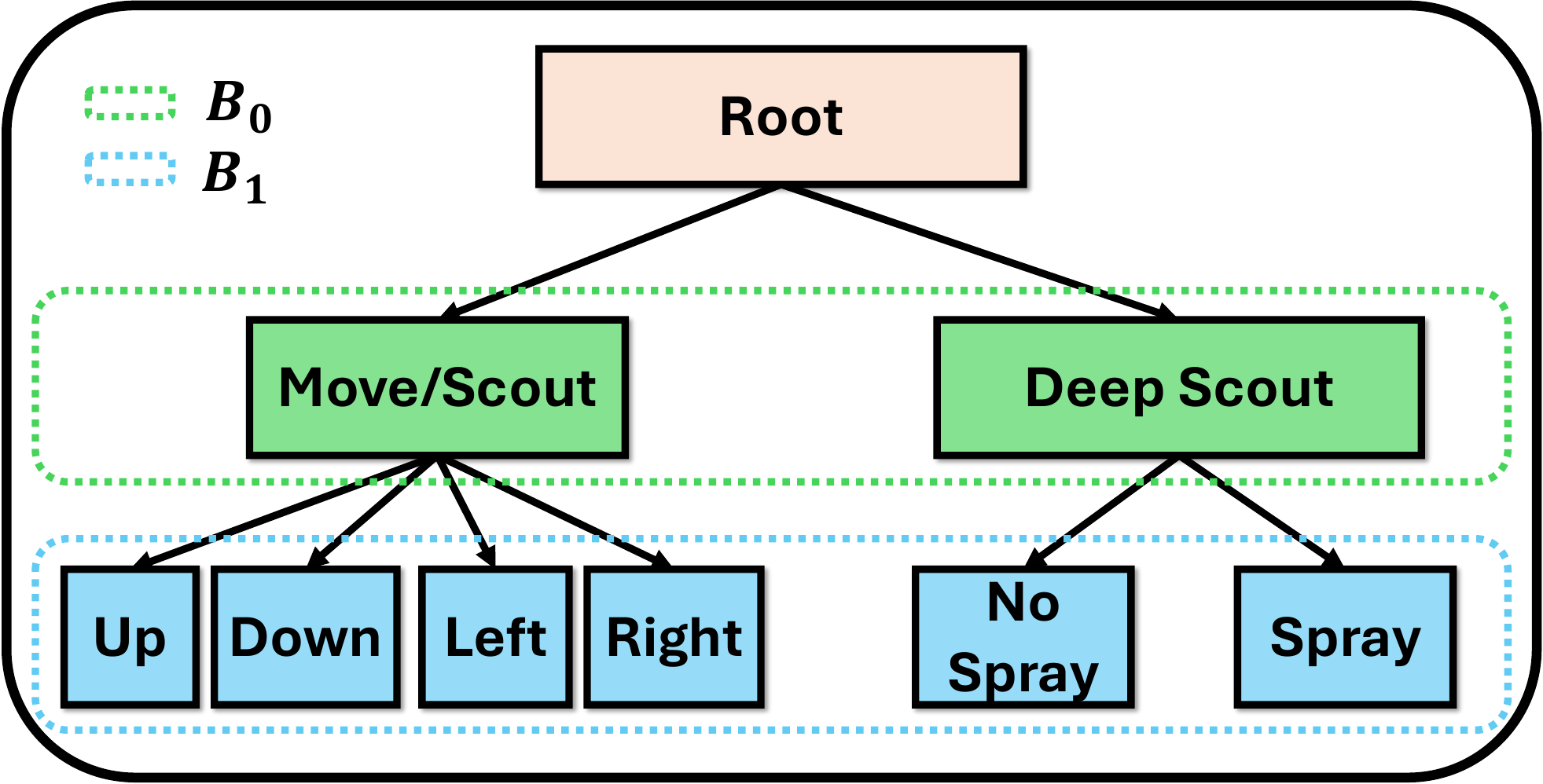}
    \caption{An action tree in our work consists of six possible actions and two components $B_0=2$ and $B_1=4\;or\;2$. The possible action is \textit{scout} or \textit{deep scout} in $B_0$, which also determines the action selected in $B_1$. Scout and deep scout cannot be performed at the same time.}
    \label{fig:action_tree}
\end{figure}
We start with defining the \textit{action trees}. Consider action $a$ as a list of finite number of components such that $a=\{b_0, b_1, ..., b_n\}$, where $n<\infty$, and $b_k, k\in\{0,1,...,n\}$ is a discrete value sampled from a set of possible values for the particular component $b_k\in B_k$. Note that actions in the same component are \textit{mutually exclusive}. We take the navigation of an agricultural robot as an example. Suppose that there are multiple actions in the component $B_k$. Hence, the possible values of $B_k$ are determined by the environment and/or robot states first, and then second by the values of the previous selections $b_0,b_1,...,b_{k-1}$. Given these restrictions, a tree structure is formed, where a path is from the root node to any leaf, which is shown in Figure~\ref{fig:action_tree}. Starting from the root node, given the current state, the robot is required to specify an action in the $B_0$ component, resulting in an action tree with a single component, $a=\{b_0\}$. In the sense of parameterized action space, $B_0$ indicates the action type while $B_1$ signifies the discrete action parameter such that the action tree can also be described by $a=\{b_0, b_1\}$, which eases the quantification of the action tree. Note that in previous work~\cite{fan2019hybrid}, the action tree is also referred to as a \textit{hierarchical action space}.
Additionally, at a particular environmental state, the agent can only perform some actions associated with a sub-tree, which is defined by leveraging the IAM. For example, in Figure~\ref{fig:action_tree}, when the robot chooses to scout, the robot may only be allowed to move up or down in the field, following the lawnmower style movement. To implement the IAM in action trees, we present the \textit{conditional action masking} (CAT) formally in the next to characterize the mechanism. 
\FloatBarrier
\begin{figure}[H]
    \centering
    \includegraphics[width=0.5\linewidth]{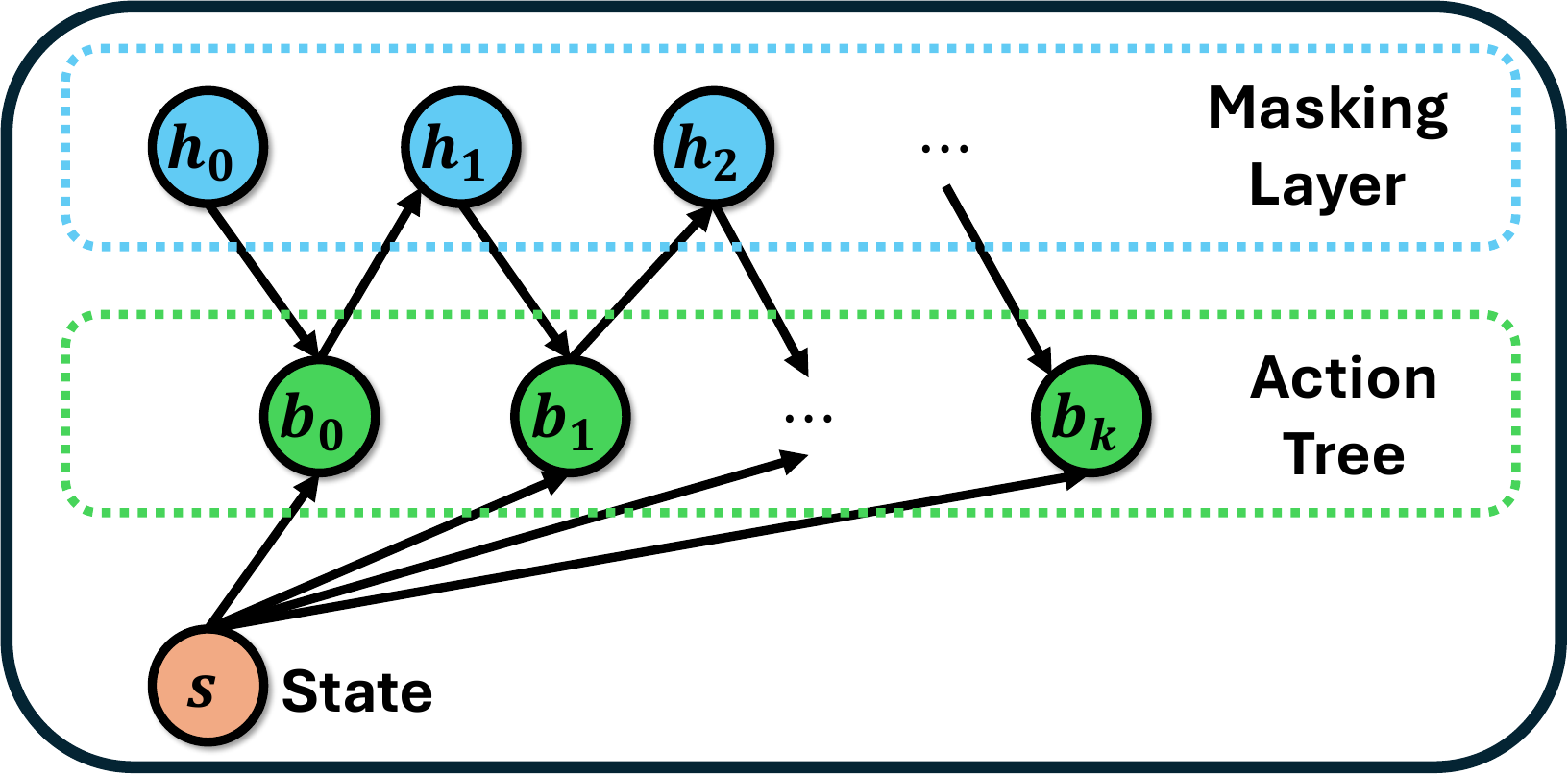}
    \caption{A schematic diagram of CAT with the joint distribution of masks $h$ and components $b$. A component $b_0$ is sampled from the options allowed by the mask $h_0$. Subsequently, $h_1$ depends on the $b_0$, determining on the next possible component $b_1$, given the state $s$. Iteratively, after all components are sampled through the CAT, the action $a$ is determined.}
    \label{fig:cat}
\end{figure}

A CAT is established by adding a mask at each node of an action tree, ensuring which child nodes of the tree are available at the present state. The action starts from the root node, which typically represents the environment state and is followed by selecting a child node from the masked distribution after applying the IAM. Denote by $h_k$ the masking function at time step $k$ such that the mask is obtained as $h_{k+1}\sim\mathbb{P}(h_{k+1}|b_k)$. With this in hand, the next component $b_{k+1}$ is sampled through the masked policy, i.e., $b_{k+1}\sim\mathbb{P}(b_{k+1}|h_{k+1},s)$. Analogously, this process will be repeated to make sure all components along the action path are sampled, as depicted in Figure~\ref{fig:cat}. Hence, the policy $\pi(a|s)$ is factorized as:
\begin{equation}
    \pi(a|s) = \mathbb{P}(h_0)\prod_{k=0}^n\mathbb{P}(b_k|h_k,s)\mathbb{P}(h_{k+1}|b_k)
\end{equation}
Figure~\ref{fig:cat} shows a generic diagram for multiple components in the action $a$. However, when the action tree is simplified to the structure as in Figure~\ref{fig:action_tree}, the policy $\pi(a|s)$ can be expressed as:
\begin{equation}
    \pi(a|s) =\mathbb{P}(h_0)\mathbb{P}(b_0|h_0,s)\mathbb{P}(h_1|b_0)\mathbb{P}(b_1|h_1,s)
\end{equation}
As the number of components are only 2, the term of $\mathbb{P}(h_2|b_1)$ to customize the next level of masking is omitted.

\section{Proposed framework}
\begin{figure}
    \centering
    \includegraphics[width=0.9\linewidth]{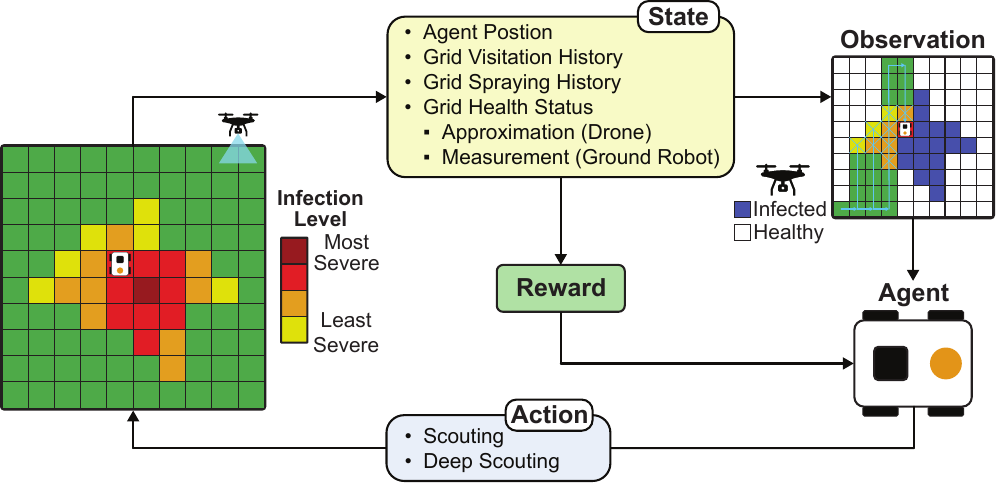}
    \caption{Schematic diagram of the proposed RL framework for precision pest management. The framework follows an iterative RL loop, where the agent observes the field state, selects high-level, and corresponding low-level decisions. Observations, including agent position, visitation history, spraying history, and plot health status (collected via drones), are updated after each action using ground robot measurements. Rewards guide the agent’s behavior, enabling it to optimize pest management strategies effectively, even in the presence of noisy or incomplete data.}
    \label{fig:framework}
\end{figure}

The proposed framework, also illustrated in Fig.~\ref{fig:framework}, outlines the agent's decision-making process utilizing a partially observable environment to optimize yield recovery and resource efficiency.

\subsection{Simulation environment}
We utilize a simulation environment, called AgGym to generate realistic infection maps for a given field. This simulator was developed in our previous work \cite{khosravi2024aggym} to model the evolution and spread of biotic stresses in agricultural fields and estimate yield losses under various pest and disease conditions. For the purpose of this paper, AgGym only provides an initial static map of infection, which serves as a representation of the health status of the agricultural field. Typically, such a map includes both healthy areas and regions with varying degrees of infection as seen in Figure~\ref{fig:framework}. Since we assume that the environment remains static during the robot's operation, we do not use the dynamic evolution of infection aspect of the AgGym simulator in this work.  

 \subsection{Overall architecture}
 The proposed architecture leverages PPO with a global actor-critic network to guide the agent’s decision-making process (see Figure~\ref{fig:framework}). The actor network has two heads: one for selecting high-level actions (scouting or deep scouting) and another for low-level actions (movement or spraying). The critic network evaluates the actions taken by the actor, providing feedback based on the estimated value of state-action pairs. The system operates within a row-crop agricultural environment, where the agent is constrained to move along designated crop rows, a typical configuration for many crops such as corn, soybeans, and cotton. The field comprises multiple grids, with each grid denoting a discrete sub-region of the field. 
Movement is restricted to row-wise traversal, prohibiting lateral or diagonal transitions except in specially designed cases; upon reaching the end of a row, the agent turns around and proceeds in the opposite direction. This constraint mirrors real-world operations of agricultural robots in row-crop fields, where equipment is typically bound by row layouts. The agent’s primary goal is to maximize yield recovery while minimizing resource consumption, including power source constraints and pesticide usage.

Furthermore, the agent’s decision-making is guided by observations of the field’s health status, which is initially noisy and imprecise, simulating real-world challenges in obtaining accurate data. In a realistic scenario, for example, we gather rough infection data through drone observations, which provide imperfect information about the field. At each step, the agent must select among feasible actions, which are constrained through IAM and CAT. This mechanism ensures that the agent only considers physically possible or strategically relevant moves based on the environment and task constraints. Within each decision cycle, the agent receives observations and chooses between two high-level actions: scouting or deep scouting.
After the agent executes its chosen action, both the state space and the observation space are updated, reflecting the outcomes of its interventions. A domain-specific reward is then assigned, accounting for yield recovery, pesticide use, and power source consumption and allowing the agent to balance pest management against resource efficiency.

If the agent’s maximum allowed steps per episode are exceeded or its pesticide budget runs out, a termination condition is triggered. These constraints ensure that the framework remains focused on improving yield recovery without incurring excessive operational costs. By iterating through these cycles—observing the field, choosing high-level actions, applying low-level decisions, and receiving feedback—the agent continually refines its strategy. In this way, the system converges on an effective balance between pest control and resource optimization, even under noisy or partial information.

\begin{algorithm}[H]  
\caption{Hierarchical Action Masking with PPO (HAM-PPO)}
\label{alg:hierarchical-action-selection}
\begin{algorithmic}[1]
\STATE \textbf{Input:} Field parameters \( l \), \( w \);  
state space $\mathcal{S} = \{s_1, s_2, s_3, s_4\}$; 
action space $\mathcal{A} = \{b_{0}, b_{1}\}$; 
infection states \(I = \{I_1, I_2, I_3\}\); 
initial power level \(E_t\)

\STATE \textbf{Initialize:} Actor network $\pi_{\theta}$, action masks $h_{0}, h_{1}$, critic network $V_{\phi}$, initial state $S_0$, and experience replay buffer $\mathcal{D}$

\FOR{each episode}
    \STATE Reset replay buffer $\mathcal{D}$
    \FOR{$t = 1$ to $T$ (time steps)}
        \STATE Apply $h_{0}$ to enforce environment-level restrictions
        \STATE Select high-level action $b_0$ from $\pi_{\text{type}}$ using the current state $S_t$
        
        \IF{$b_0 = \text{scouting}$}
            \STATE Apply $h_{1}$ to enforce movement constraints
            \STATE Select low-level action $b_1$ from $\pi_{\text{move}}$
        \ELSE
            \STATE Select low-level action $b_1$ from $\pi_{\text{spray}}$
        \ENDIF
        
        \STATE Execute action $a_t = (b_0, b_1)$ in the environment
        \STATE Receive next observation $o_{t+1}$, reward $r_t$, and done flag $d_t$
        \STATE Store experience tuple $(o_t, b_0, b_1, r_t, o_{t+1}, d_t)$ in replay buffer $\mathcal{D}$
        
        \IF{$d_t = \text{True}$}
            \STATE Reset environment: $o_{t+1} = \text{env.reset()}$
        \ENDIF
    \ENDFOR
    \STATE Update $\pi_{\theta}$ using PPO-Clip (\text{See Algorithm \ref{alg:ppo-clip}} for PPO-Clip update rules) with advantage estimates computed via GAE
\ENDFOR

\end{algorithmic}
\end{algorithm}
 \subsection{Algorithm}
 In this section, we describe the hierarchical action selection algorithm with action masking used in our approach (see Algirithm \ref{alg:hierarchical-action-selection}). 
 The agent's decision-making is driven by a single global actor network, $\pi_{\theta}$, which has two heads for selecting high-level and low-level actions. The high-level action policy, $\pi_{\text{type}}$, determines the broad strategy, such as whether the agent should perform ``scouting" or ``deep scouting". After the high-level action is chosen, a shared policy network selects the corresponding low-level action. To ensure that the agent only selects valid actions, the algorithm applies action masking using masks $h_{0} \text{ and } h_{1}$. Please note that $h_0$ and $h_1$ are determined by domain knowledge and not learnable. $h_{0}$ enforces environment-level restrictions, such as the prohibition of certain actions at the edges of a row-crop field or the prohibition of diagonal movement in strictly row-based navigation. Once the high-level action is chosen, we apply the corresponding low-level mask $h_{1}$. In other words, $h_{1}$ ensures that only valid movement or spraying actions are available, depending on the high-level strategy. By layering these two masks, we ensure that, at each step, the agent can only pick actions that are both environmentally permissible and consistent with its selected high-level approach.
 
 At each time step, the agent records an experience tuple 
\((o_t,\, a_{\text{type}},\, a_t,\, r_t,\, o_{t+1},\, d_t)\) 
in a replay buffer \(D\). Here, \(d_t\) is a boolean flag indicating whether the episode has terminated. At the end of each episode, the policy is updated via PPO-Clip (see Algorithm \ref{alg:ppo-clip}), which uses advantage estimates computed via GAE to stabilize learning. By iterating through this cycle of observation, action selection, and PPO-Clip updates, the agent refines its policy to reduce pesticide usage while maximizing yield recovery. We refer to this overall approach as Hierarchical Action Masking with PPO (HAM-PPO) for convenience.

\begin{algorithm}[H]
    \caption{PPO-Clip}
    \label{alg:ppo-clip}
\begin{algorithmic}[1]
    \STATE \textbf{Input:} initial policy parameters $\theta_0$, initial value function parameters $\phi_0$
    \FOR{$k = 0,1,2,...$}
    \STATE Collect set of trajectories ${\mathcal D}_k = \{\tau_i\}$ by running policy $\pi_k = \pi(\theta_k)$ in the environment.
    \STATE Compute rewards-to-go $\hat{R}_t$.
     \STATE Compute advantage estimates, $\hat{\mathbf{A}}^{\pi_{\theta_k}}$, using GAE based on the current value function $V_{\phi_k}$
    \STATE Update the policy by maximizing the PPO-Clip objective:
        \begin{equation*}
        \theta_{k+1} = \arg \max_{\theta} \frac{1}{|{\mathcal D}_k| T} \sum_{\tau \in {\mathcal D}_k} \sum_{t=0}^T \min\left(
            \frac{\pi_{\theta}(a_t|s_t)}{\pi_{\theta_k}(a_t|s_t)}  \hat{\mathbf{A}}^{\pi_{\theta_k}}(s_t,a_t), \;\;
            g(\epsilon, \hat{\mathbf{A}}^{\pi_{\theta_k}}(s_t,a_t))
        \right),
        \end{equation*}
        where
\begin{equation*}
g(\epsilon, \hat{\mathbf{A}}^{\pi_{\theta_k}}) = \text{clip}\left( \frac{\pi_{\theta}(a_t|s_t)}{\pi_{\theta_k}(a_t|s_t)}, 1 - \epsilon, 1 + \epsilon \right) \cdot \hat{\mathbf{A}}^{\pi_{\theta_k}}.
\end{equation*}
        typically via stochastic gradient ascent with Adam.
    \STATE Fit value function by regression on mean-squared error:
        \begin{equation*}
        \phi_{k+1} = \arg \min_{\phi} \frac{1}{|{\mathcal D}_k| T} \sum_{\tau \in {\mathcal D}_k} \sum_{t=0}^T\left( V_{\phi} (s_t) - \hat{R}_t \right)^2,
        \end{equation*}
        typically via stochastic gradient ascent with Adam.
    \ENDFOR
\end{algorithmic}
\end{algorithm}

\section{Experimental Setup}
This section outlines the experiments designed to evaluate the performance of the proposed planning algorithm and assess its effectiveness in optimizing decision-making and resource allocation. To simulate realistic agricultural environments, we considered the following key aspects in our experimental setup:
\subsection{Environment settings}
We model the spread of infection across the agricultural field by considering three distinct levels of pest infestation. These infection levels range from 1 to 3, and their severity is determined by the infection duration, which corresponds to the number of days that have passed since the initial onset of infection. The infection duration plays a crucial role in the yield loss, as it directly impacts the severity of the damage to the crops as defined in the yield loss Eq. 
\eqref{eq:yield_loss}. The agent moves through the field using scouting, which takes \( \tau_s = 1 \) time step  and consumes minimal power. Additionally, the agent can perform deep scouting, which takes \( \tau_d = 5 \) time steps and requires higher energy consumption due to its detailed assessment process. To encourage efficient exploration and prevent unnecessary revisits, a penalty of \( \kappa_h = -5 \) is applied whenever the agent revisits a previously scouted healthy crop. The value of -5 is based on some manual tuning efforts given the problem.

\subsection{Observation noise modeling}
In real-world agricultural applications, partial observability is a common challenge due to the inherent limitations of drone-based infection detection. Factors such as weather conditions, sensor inaccuracies, and environmental interference can introduce uncertainty into the collected infection data. Before the robot navigates through the field, rough infection data is typically gathered through drone observations, which may be imperfect and noisy.

To simulate this uncertainty, we model the infection status of each crop unit as a binary value, indicating whether it is infected (1) or healthy (0). However, to reflect real-world imprecision, we introduce two types of observation noise:
\begin{enumerate}
    \item \textbf{Gaussian (Uniform) Noise:} This noise mimics gradual distortions in the observed infection status by adding perturbations to the actual health values (0 for healthy, 1 for infected). Specifically, we model the uncertainty as \( \nu \sim \mathcal{N}(0, \sigma^2) \), where \( \sigma \) determines the level of noise. This formulation captures continuous variations in observation accuracy, which are common in noisy sensor readings, and accurately reflects real-world noise in many systems~\cite{bulsara1996tuning}.

    
    \item \textbf{Binary Flip Noise:} This noise simulates abrupt misclassifications by randomly flipping the binary infection labels, meaning that a healthy crop (0) can be misclassified as infected (1) and vice versa. This type of noise accounts for discrete errors that may occur due to occlusions, poor lighting conditions, or sensor failures.
\end{enumerate}

\subsection{Baseline methods}


We compare the performance of our proposed algorithm, HAM-PPO, with various baseline approaches across different environmental scenarios. These scenarios help assess the generalizability and adaptability of our algorithm compared to simpler movement and spraying strategies and are used for both comparative evaluation and ablation studies. The following baseline policies are considered:

\noindent\textbf{Lawnmower Movement Pattern with Carpet Spraying:} This baseline assumes the agent follows a predefined path across the field in a systematic pattern (similar to mowing a lawn) while applying pesticide uniformly to all grids it passes. This approach represents a traditional, non-adaptive method in agricultural robots for coverage.

\noindent\textbf{The Lawnmower Movement Pattern with Reactive Spraying:} This baseline emulates a sensor-based, targeted intervention approach in which the robot follows a predetermined, systematic lawnmower pattern for comprehensive field coverage. Simultaneously, a visual perception module—employing computer vision techniques— monitors  crop conditions to identify potential infection. When the system detects an anomaly suggestive of infection, it immediately triggers the spraying mechanism in that localized area. However, due to inherent uncertainties in the visual detection process—such as environmental variability, sensor noise, or suboptimal image resolution—the algorithm may sometimes misclassify healthy areas as infected or overlook actual infections. To simulate sensor uncertainty, we incorporate an observation flip noise model. In this model, binary indicators (0 for healthy, 1 for infected) are randomly flipped with a specified probability (0.15), which can result in false positives (healthy misclassified as infected) and false negatives (infected misclassified as healthy).

\noindent\textbf{Lawnmower Movement Pattern with Optimal Spraying:} This baseline also uses the lawnmower pattern for movement but incorporates our proposed policy for targeted spraying. The agent adapts its spraying strategy based on observations of the field’s health status, optimizing pesticide use by spraying only the affected grids.

\noindent\textbf{Random Policy:} This baseline introduces a purely random approach where the agent selects actions arbitrarily. It randomly chooses between high-level actions (scouting or deep scouting) and low-level actions (moving in one of four directions or choosing to spray or not spray). This represents a completely uninformed decision-making strategy, providing a stark contrast to the more strategic approaches of HAM-PPO.

\subsection{Comparative evaluation}
For each baseline method, we conduct evaluations using the same environmental scenarios as those applied to our proposed algorithm (represented in Tables~\ref{table:scenarios-yieldpercentage}--\ref{table:scenarios-pesticidecost} as Optimal Navigation + Optimal Spray (ours)). The scenarios characterized by varied infection ranges, randomness in infection spread, and infection initiation locations, as detailed in  Tables~\ref{table:scenarios-yieldpercentage}--\ref{table:scenarios-pesticidecost}. The battery budget represents the available power resources, which determine the maximum number of timesteps the agent can operate. To introduce partial observability into the evaluation, we incorporate Gaussian noise (with a mean of 0 and a standard deviation of 0.15) into the infection data. Initially, a global policy is trained to handle all environmental scenarios, including those with corner infection initiation. Afterward, this global policy is fine-tuned for center infection initiation scenarios. The fine-tuned policy is then evaluated across the center infection initiation scenarios. Each scenario is tested using five different random seeds to ensure robustness and generalizability of the results.
The scenarios tested include varying infection ranges, considering (20-30\%) and (30-40\%) infection percentages within the field. Next, we assess the impact of infection randomness and spatial distribution, testing two categories: \textit{moderate randomness}, where infection spreads with some spatial clustering, and \textit{low randomness}, where the spread follows a deterministic and predictable pattern. We also explore two infection distribution types: Categorical Infection Distribution, with infections spread at three severity levels (high, moderate, low), and Fixed Infection Distribution, where infection spread is deterministic and predefined. Finally, we examine the effect of different infection initiation locations: starting from the center of the field, or from two corners, simulating different spread patterns.  

The scenarios tested include varying infection ranges, considering (20-30\%) and (30-40\%) infection 
percentages within the field. The battery budget represents the available power resources, 
which determine the maximum number of timesteps the agent can operate. 

To establish an appropriate timestep budget for each infection range, we account for both movement 
and deep scouting (spraying) actions. Given a \(10 \times 10\) field, a 20-30\% infection rate corresponds to an average of 25 infected grids, while a 30-40\% infection rate corresponds to 35 infected grids. 
Each deep scouting action requires 5 timesteps, meaning that for the 20-30\% infection scenario, 
spraying alone consumes:

\[
25 \times 5 = 125 \text{ timesteps}
\]

In addition, to navigate the field and locate infected areas, the robot must traverse multiple rows and columns, adjusting its path based on the spatial distribution of infections. Depending on the infection spread, it may either move continuously down a column before switching to the next or adjust mid-way if a 
neighboring infected grid allows for a shorter traversal. However, in both cases, the number of 
grids covered remains approximately the same.

To estimate movement cost, we consider that the robot must cover a number of rows and columns 
comparable to structured traversal methods, but in a more adaptive way. The movement cost consists of: 
Additional buffer to ensure reasonable coverage of infections, accounting for spatial uncertainty and infection clustering:
\begin{itemize}
    \item 50 timesteps for the lower infection range (20-30\%).
    \item 60 timesteps for the higher infection range (30-40\%).
\end{itemize}

Thus, the total movement cost is:
\[
M = 50 + 5 + 50 = 105 \quad \text{(for 20-30\% infection)}
\]
\[
M = 50 + 5 + 60 = 115 \quad \text{(for 30-40\% infection)}
\]

The total timestep budget is then computed as:
\[
T = 105 + (25 \times 5) = 230 \quad \text{timesteps (for 20-30\%)}
\]
\[
T = 115 + (35 \times 5) = 290 \quad \text{timesteps (for 30-40\%)}
\]

This explicit buffer ensures that the agent has sufficient flexibility to navigate effectively while accounting for infections in neighboring areas.







\subsection{Training details}
We implement HAM-PPO, building upon the PPO masking technique from stable-baselines3 outlined in the documentation \cite{ppo-mask}. We closely follow the hyperparameter setup from the documentation as they have been tuned nearly-optimal. The model is trained for a total of 6 million steps to ensure sufficient exploration and learning across varying scenarios for the global policy. Additionally, for each optimized policy tailored to specific scenarios, used to compare against the global policy, fine-tuning is conducted for 3 million steps per scenario. The details of the hyperparameters and network structure used in our experiments are provided in Table~\ref{tab:ppo_hyperparams}.

\begin{table}[h]
    \centering
    \caption{Hyperparameters used for PPO-Mask.}
    \label{tab:ppo_hyperparams}
    \begin{tabular}{ll}
        \toprule
        \textbf{Hyperparameter} & \textbf{Value} \\
        \midrule
        Optimizer & Adam \\
        Learning rate & $3 \times 10^{-4}$ \\
        Update interval (n\_steps) & 2048 \\
        Minibatch size & 64 \\
        Number of epochs & 10 \\
        Discount factor ($\gamma$) & 0.99 \\
        GAE parameter ($\lambda$) & 0.95 \\
        Clipping parameter ($\epsilon$) & 0.2 \\
        Entropy coefficient ($c_{ent}$) & 0.02 \\
        Value function coefficient ($c_{vf}$) & 0.5 \\
        Gradient norm clipping & 0.5 \\
        Activation function & ReLU \\
        Policy network ($\pi$) & 256 → 128 → 128 → 64 \\
        Value function ($v_f$) & 256 → 128 → 128 → 64 \\
        \bottomrule
    \end{tabular}
\end{table}

\section{Results and discussion}
\subsubsection{Comparative Evaluation}
This subsection compares HAM-PPO’s performance against baseline methods across different environmental conditions.  These scenarios help assess the generalizability and adaptability of our algorithm compared to simpler movement and spraying strategies. The results focus on key performance metrics, including yield recovery, economic efficiency, and pesticide cost, to demonstrate the effectiveness of HAM-PPO in optimizing pesticide application and resource allocation.

\begin{table}[h!]
\small
\caption{Comparison of Models on \textit{Yield Recovered} with Different Environmental Parameters }
\label{table:scenarios-yieldpercentage}
\resizebox{\textwidth}{!}{  
\begin{tabular}{|c|c|c|c|c|c|c|c|c|}
\hline
\multirow{4}{*}{\textbf{Model used}}                                    & \multicolumn{8}{c|}{\textbf{Environmnet   Parameter}} \\ \cline{2-9}
        & \multicolumn{4}{c|}{20-30\%(infected),   Battery budget=230}                                                                                                                                                                                                                    & \multicolumn{4}{c|}{30-40\%(infected), Battery   Budget=290}   \\ \cline{2-9}
   & \multicolumn{2}{c|}{\begin{tabular}[c]{@{}c@{}}Low\\       Randomness\end{tabular}}                                                   & \multicolumn{2}{c|}{\begin{tabular}[c]{@{}c@{}}High\\       Randomness\end{tabular}}                                                     & \multicolumn{2}{c|}{\begin{tabular}[c]{@{}c@{}}Low\\       Randomness\end{tabular}}                                                    & \multicolumn{2}{c|}{\begin{tabular}[c]{@{}c@{}}High\\       Randomness\end{tabular}}                                                     \\ \cline{2-9}
  & \begin{tabular}[c]{@{}c@{}}center\\       infection\end{tabular} & \begin{tabular}[c]{@{}c@{}}conrner\\       infection\end{tabular} & \begin{tabular}[c]{@{}c@{}}center\\       infection\end{tabular} & \begin{tabular}[c]{@{}c@{}}conrners   \\      infection\end{tabular} & \begin{tabular}[c]{@{}c@{}}center\\       infection\end{tabular} & \begin{tabular}[c]{@{}c@{}}conrners\\       infection\end{tabular} & \begin{tabular}[c]{@{}c@{}}center   \\      infection\end{tabular} & \begin{tabular}[c]{@{}c@{}}conrners\\       infection\end{tabular} \\ \hline
\multicolumn{9}{|c|}{\textbf{Yield recovered (\%)}}\\ \hline
\begin{tabular}[c]{@{}c@{}} Optimal Navigation + \\     Optimal Spray (HAM-PPO, ours)
\end{tabular}& 79                                                             & 84                                                             & 76                                                             & 82                                                                 & 77                                                             & 75                                                               & 72                                                               & 73                                                               \\ \hline
\begin{tabular}[c]{@{}c@{}}LawnMower+\\      Optimal Spray (ours) \end{tabular} &75                                                                  &  76                                                                  & 72                                                                  & 77                                                                     &74                                                                  &70                                                                    & 74                                                                  &76                                                                    \\ \hline
\begin{tabular}[c]{@{}c@{}}LawnMower+\\      Reactive Spray \end{tabular} &62
&68
&55
&63
&65
&67
&62
&69

\\ \hline
\begin{tabular}[c]{@{}c@{}}LawnMower+\\      Carpet Spary\end{tabular}  & 52                                                            & 62                                                            & 50                                                            & 48                                                                & 66                                                            & 43                                                              & 59                                                               & 56                                                              \\ \hline
RandomPolicy                                                            & 1                                                              & 4                                                              & 1                                                             & 8                                                                 & 10                                                             & 15                                                               & 15                                                              & 14   \\ \hline                                                           
\end{tabular}
}

\end{table}

\begin{table}[h!]
\small
\caption{Comparison of Models on \textit{Yield Recovered, Bushel Per Acre (\$)} with Different Environmental Parameters}
\label{table:scenarios-yieldprice}
\resizebox{\textwidth}{!}{  
\begin{tabular}{|c|c|c|c|c|c|c|c|c|}
\hline
\multirow{4}{*}{\textbf{Model used}}                                    & \multicolumn{8}{c|}{\textbf{Environmnet   Parameter}} \\ \cline{2-9}
        & \multicolumn{4}{c|}{20-30\%(infected),   Battery budget=230}                                                                                                                                                                                                                    & \multicolumn{4}{c|}{30-40\%(infected), Battery   Budget=290}   \\ \cline{2-9}
   & \multicolumn{2}{c|}{\begin{tabular}[c]{@{}c@{}}Low\\       Randomness\end{tabular}}                                                   & \multicolumn{2}{c|}{\begin{tabular}[c]{@{}c@{}}High\\       Randomness\end{tabular}}                                                     & \multicolumn{2}{c|}{\begin{tabular}[c]{@{}c@{}}Low\\       Randomness\end{tabular}}                                                    & \multicolumn{2}{c|}{\begin{tabular}[c]{@{}c@{}}High\\       Randomness\end{tabular}}                                                     \\ \cline{2-9}
  & \begin{tabular}[c]{@{}c@{}}center\\       infection\end{tabular} & \begin{tabular}[c]{@{}c@{}}conrner\\       infection\end{tabular} & \begin{tabular}[c]{@{}c@{}}center\\       infection\end{tabular} & \begin{tabular}[c]{@{}c@{}}conrners   \\      infection\end{tabular} & \begin{tabular}[c]{@{}c@{}}center\\       infection\end{tabular} & \begin{tabular}[c]{@{}c@{}}conrners\\       infection\end{tabular} & \begin{tabular}[c]{@{}c@{}}center   \\      infection\end{tabular} & \begin{tabular}[c]{@{}c@{}}conrners\\       infection\end{tabular} \\ \hline
\multicolumn{9}{|c|}{\textbf{Yield recovered, Bushel per acre (\$)}}\\ \hline
\begin{tabular}[c]{@{}c@{}} Optimal Navigation + \\     Optimal Spray (HAM-PPO, ours)
\end{tabular}                                                  & 51                              & 42                              & 46                            & 44                           & 53                            & 57                             & 50                            & 57                                      \\ \hline
\begin{tabular}[c]{@{}c@{}}LawnMower+\\      Optimal Spray (ours) \end{tabular} & 44                                   &39                                    & 40                                   &42                                    & 50                                   & 53                                   & 51                                   &59                                                                                                       \\ \hline
\begin{tabular}[c]{@{}c@{}}LawnMower+\\      Reactive Spray \end{tabular} &33
&38
&30
&38
&46
&55
&48
&45
\\ \hline
\begin{tabular}[c]{@{}c@{}}LawnMower+\\      Carpet Spary\end{tabular}   & 32                              & 32                              & 32                             & 27                              & 48                             & 46                             & 40                           & 29                                                                \\ \hline
RandomPolicy                                                             & 6                               & 6                               & 6                              & 8                             & 6                             & 10                             & 10                             & 10    \\ \hline

\end{tabular}
}

\end{table}

\begin{table}[h!]
\small
\caption{Comparison of Models on \textit{Pesticide Cost (\$) Per Acre} with Different Environmental Parameters}
\label{table:scenarios-pesticidecost}
\resizebox{\textwidth}{!}{  
\begin{tabular}{|c|c|c|c|c|c|c|c|c|}
\hline
\multirow{4}{*}{\textbf{Model used}}                                    & \multicolumn{8}{c|}{\textbf{Environmnet   Parameter}} \\ \cline{2-9}
        & \multicolumn{4}{c|}{20-30\%(infected),   Battery budget=230}                                                                                                                                                                                                                    & \multicolumn{4}{c|}{30-40\%(infected), Battery   Budget=290}   \\ \cline{2-9}
   & \multicolumn{2}{c|}{\begin{tabular}[c]{@{}c@{}}Low\\       Randomness\end{tabular}}                                                   & \multicolumn{2}{c|}{\begin{tabular}[c]{@{}c@{}}High\\       Randomness\end{tabular}}                                                     & \multicolumn{2}{c|}{\begin{tabular}[c]{@{}c@{}}Low\\       Randomness\end{tabular}}                                                    & \multicolumn{2}{c|}{\begin{tabular}[c]{@{}c@{}}High\\       Randomness\end{tabular}}                                                     \\ \cline{2-9}
  & \begin{tabular}[c]{@{}c@{}}center\\       infection\end{tabular} & \begin{tabular}[c]{@{}c@{}}conrner\\       infection\end{tabular} & \begin{tabular}[c]{@{}c@{}}center\\       infection\end{tabular} & \begin{tabular}[c]{@{}c@{}}conrners   \\      infection\end{tabular} & \begin{tabular}[c]{@{}c@{}}center\\       infection\end{tabular} & \begin{tabular}[c]{@{}c@{}}conrners\\       infection\end{tabular} & \begin{tabular}[c]{@{}c@{}}center   \\      infection\end{tabular} & \begin{tabular}[c]{@{}c@{}}conrners\\       infection\end{tabular} \\ \hline
\multicolumn{9}{|c|}{\textbf{Pesticide cost (\$) per acre}}\\ \hline
\begin{tabular}[c]{@{}c@{}} Optimal Navigation + \\     Optimal Spray (HAM-PPO, ours)
\end{tabular}                                                                 & 1.14                             & 0.84                            & 1.17                             & 0.93                             & 1.23                             & 0.96                              & 1.2                             & 1                                                           \\ \hline
\begin{tabular}[c]{@{}c@{}}LawnMower+\\     Optimal Spray (ours)\end{tabular} &1.17                                    & 0.84                                   &1.23                                    &0.93                                    & 1.23                                   &0.96                                    & 1.2                                   &1.11    \\ \hline
\begin{tabular}[c]{@{}c@{}}LawnMower+\\      Reactive Spray \end{tabular}&0.75
&0.81
&0.6
&0.75
&0.83
&0.85
&0.82
&0.88
   \\ \hline  
\begin{tabular}[c]{@{}c@{}}LawnMower+\\      Carpet Spary\end{tabular}   & 1.3                              & 1.3                              & 1.3                             & 1.3                             & 1.41                              & 1.41                              & 1.41                              & 1.41                                            \\ \hline
RandomPolicy                                                           & 0.57                               & 0.54                               & 0.54                              & 0.57                               & 0.63                              & 0.63                              & 0.6                              & 0.6                              \\ \hline                                                           
\end{tabular}
}

\end{table}

\noindent\textbf{Yield Recovered (\%).}
    The Yield Recovered metric, expressed as a percentage, measures the proportion of the initially infected units (or grids) in the field that have successfully been sprayed and recovered. 
    From the results presented in Table \ref{table:scenarios-yieldpercentage}, we observe that HAM-PPO outperforms all baseline models across all scenarios, consistently achieving higher recovery percentages. For example, in the scenario of 20-30\% infection with a Battery Budget of 230, HAM-PPO achieves a yield recovery of 79\% under low randomness and 84\% under high randomness, which is significantly higher than LawnMower+Carpet Spray (52\% and 62\%) and RandomPolicy (1\% and 4\%).
    Notably, LawnMower+Optimal Spray performs better than LawnMower+Carpet Spray, showing that combining movement strategies with HAM-PPO’s spraying policy improves recovery. Additionally, the LawnMower+Reactive Spray baseline attains intermediate recovery rates around 62–68\% under the same conditions. Its performance reflects the trade-off between the comprehensive coverage of a lawnmower pattern and the inherent detection errors from its sensor-driven approach (modeled via observation flip noise), especially under conditions of higher randomness or corner infection initiation. The RandomPolicy remains the least effective, with yield recoveries consistently low across all scenarios. LawnMower+Optimal Spray exhibits competitive performance, particularly in scenarios with 30-40\% infection and a Battery Budget of 290. In high randomness and corner infection settings, it achieves yield recoveries of up to 76\%, slightly outperforming HAM-PPO's 73\%. This slight advantage is expected, as LawnMower’s systematic movement strategy ensures comprehensive coverage in sparser and more unpredictable infection distributions, where HAM-PPO’s adaptive movement may face challenges in covering the field evenly.

\noindent\textbf{Yield Recovered (\$) per Bushel.}
    The Yield Recovered (\$) per Bushel metric translates the yield recovery into monetary terms. It is calculated by determining the total number of recovered grids in a $100\times 100$ square foot field, where each grid represents a unit of filed. The attainable yield per grid is multiplied by the price per bushel, then scaled by the total number of recovered units. This metric helps assess the financial benefits of applying the spraying policies. As shown in Table \ref{table:scenarios-yieldprice}, HAM-PPO again demonstrates superior financial performance, with the highest recovery values across all scenarios. For instance, in the 20-30\% infection scenario with Battery Budget 230, HAM-PPO achieves a recovery value of \$51 per bushel, compared to \$32 for LawnMower+Carpet Spray and \$6 for RandomPolicy. The LawnMower+Optimal Spray policy further improves performance, with values ranging from \$40 to \$56 across different scenarios, and reaching up to \$59 in challenging conditions with high randomness and corner infections—slightly outperforming HAM-PPO (\$57). Meanwhile, the LawnMower+Reactive Spray baseline delivers moderate economic returns. In the same 20–30\% infection scenario with Battery Budget 230, it records recovery values around \$33–\$38 per bushel. This moderate performance reflects the inherent trade-off between broad coverage and detection accuracy. 
    

\noindent\textbf{Pesticide Cost (\$) per Acre.}
    The Pesticide Cost (\$) per Acre metric indicates the financial cost of pesticide usage in the field. It reflects how efficient the spraying policies are in terms of their operational cost, with lower costs being preferable. In the Pesticide Cost (\$) per Acre metric (Table \ref{table:scenarios-pesticidecost}), HAM-PPO achieves a balanced performance; for example, in the 20–30\% infection scenario with a Battery Budget of 230, HAM-PPO incurs a cost of \$1.14 per acre, which is lower than that of LawnMower+Carpet Spray at \$1.29, while the RandomPolicy records the lowest cost due to its minimal pesticide usage, albeit at the expense of yield recovery.
    
    Notably, the LawnMower+Reactive Spray baseline generally incurs the lowest pesticide costs across several scenarios, recording approximately \$0.75–\$0.81 per acre in the 20–30\% infection scenario with a Battery Budget of 230. This cost efficiency stems from its targeted reactive spraying approach. However, these savings must be weighed against its lower yield recovery, as sensor uncertainty (modeled by observation flip noise) can cause misclassification and missed infections, resulting in reduced overall recovery.

\begin{figure}
    \centering
    \includegraphics[width=1.05\linewidth]{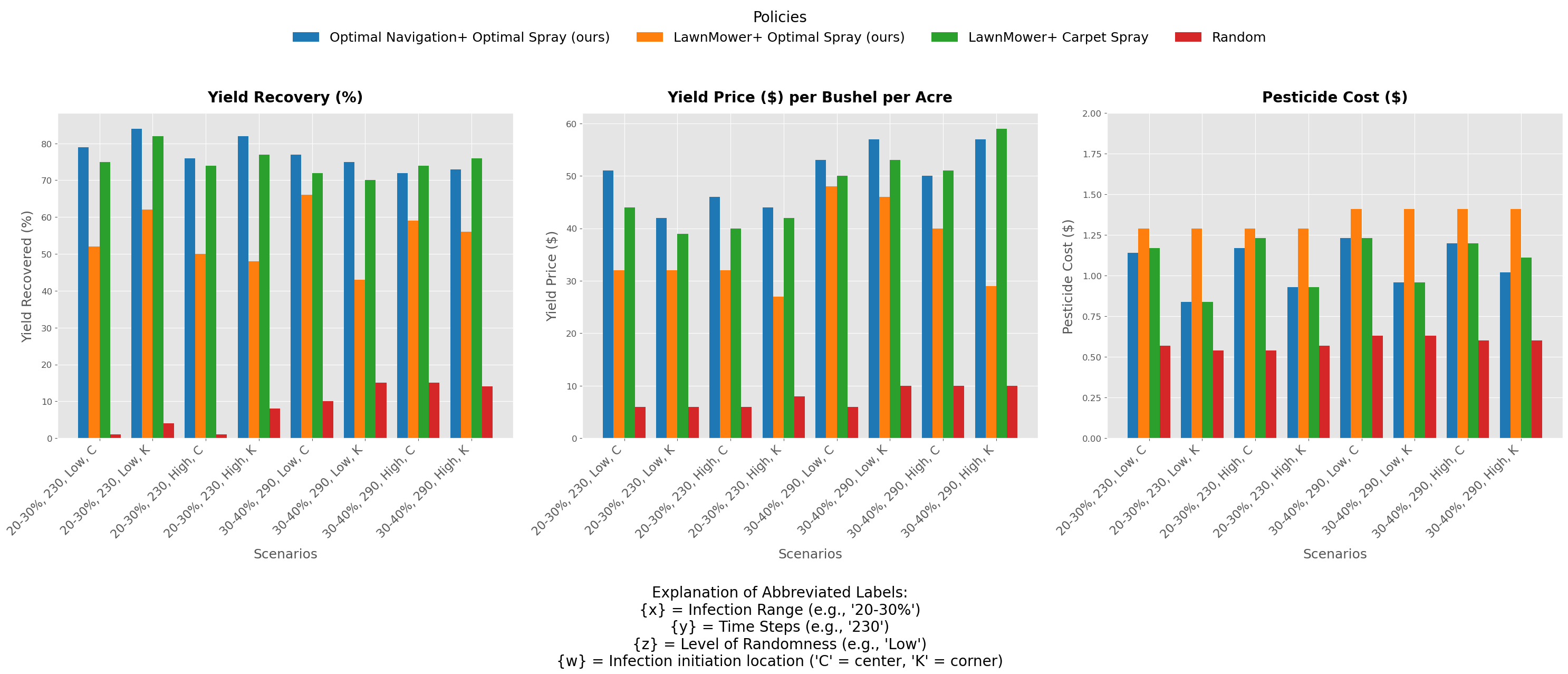}
    \caption{Each subplot evaluates a distinct performance metric: (a) Yield Recovery (\%) measures the percentage of yield preserved from potential losses, (b) Yield Price (\$) per Bushel per Acre reflects the economic value of recovered yield per unit area, and (c) Pesticide Cost (\$) quantifies the expenses associated with pesticide application.}
    \label{fig:comparative}
\end{figure}

These findings are further illustrated in Figure~\ref{fig:comparative}, where the line plots provide a visual representation of performance across all scenarios and metrics. The figure highlights the consistent advantages of HAM-PPO’s spraying policy in achieving efficient and targeted pesticide application. While LawnMower+HAM-PPO Spray shows comparable results in some scenarios,  its performance further highlights the effectiveness of HAM-PPO's efficient spraying policy, particularly when compared to LawnMower+Carpet Spray. In contrast, HAM-PPO’s adaptive decision-making consistently demonstrates robust performance, particularly in reducing pesticide usage and enhancing recovery across diverse environmental conditions.

\subsection{Ablation studies}
The following ablation studies were conducted to evaluate the performance of our proposed HAM-PPO algorithm across different experimental settings:

\noindent\textbf{Impact of Different Amounts of Observation Noise on Model Performance:} 

In this study, we investigated how varying levels of noise in the observation space affect our model's performance by evaluating two types of noise: Gaussian noise, which introduces gradual distortions with varying standard deviations \( \sigma \in \{0.05, 0.2, 0.5, 0.7\} \), and binary flip noise, which simulates abrupt misclassifications by flipping infection labels with probabilities \( p \in \{0.05, 0.2, 0.5, 0.7\} \) in drone-captured observations.

\begin{figure}
    \centering
    \includegraphics[width=1.0\linewidth]{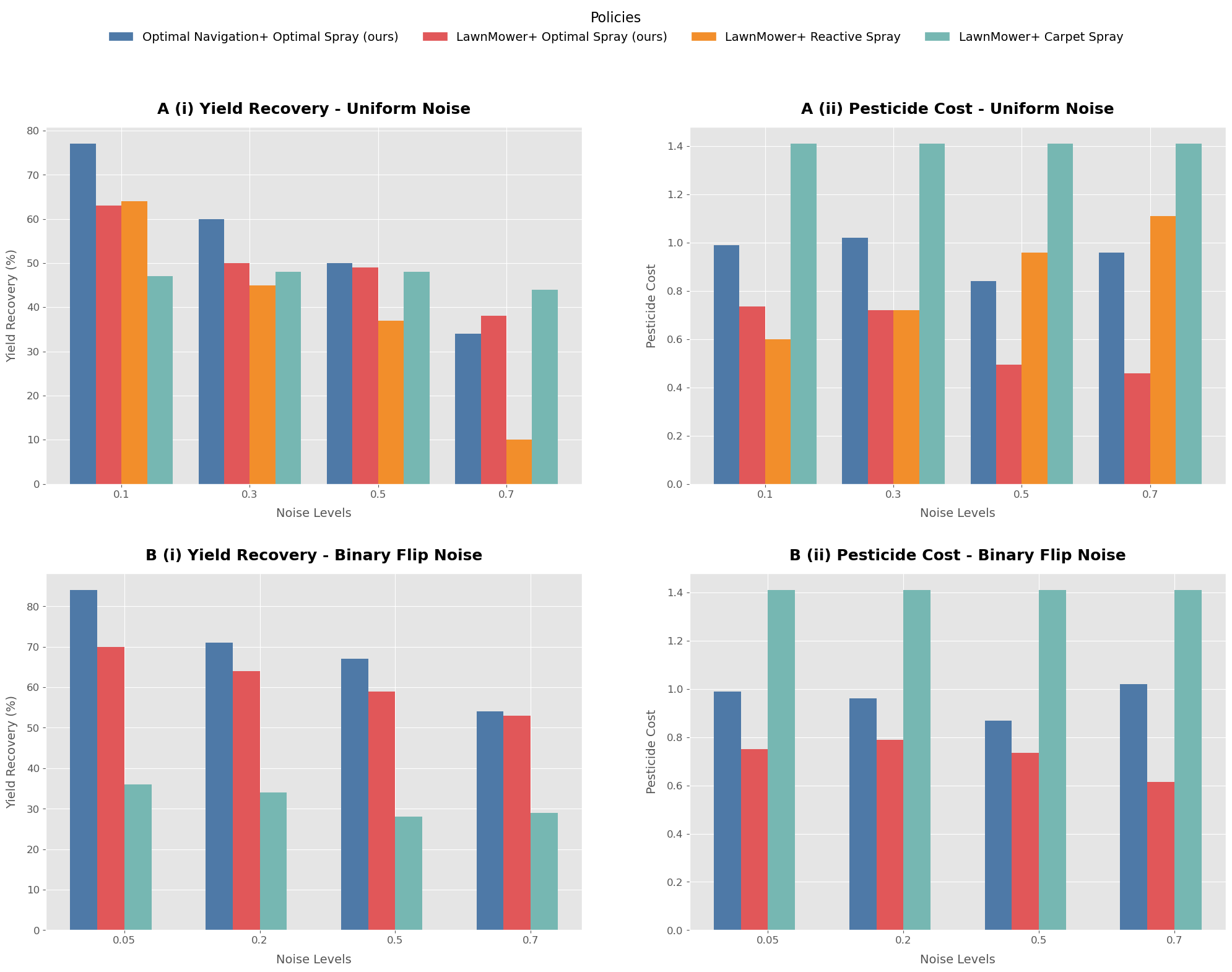}
    \caption{Impact of observation noise on the performance of HAM-PPO. (A) illustrates the effects of Gaussian noise (with \(\sigma\) ranging from 0.05 to 0.7) on yield recovery (i) and pesticide cost (ii). (B) shows the corresponding impact of binary flip noise (with flipping probability \(p \in \{0.05, 0.2, 0.5, 0.7\}\)) on these metrics. }
    \label{fig:noiselevel}
\end{figure}

Figures~\ref{fig:noiselevel}A and~\ref{fig:noiselevel}B summarize the results. As the noise level increases—whether through larger $\sigma$ or higher flipping probability—performance degrades for all methods. However, Optimal Navigation+Optimal Spray (HAM-PPO, ours) consistently shows the highest yield recovery under both noise types, reflecting its resilience to observation uncertainty.  
Meanwhile, LawnMower+Reactive Spray tends to maintain a relatively low pesticide cost due to its targeted approach but suffers more from misclassifications, leading to reduced yield recovery at higher noise levels. These findings highlight that a carefully designed policy (such as our hierarchical approach) can better tolerate inaccuracies in infection indicators compared to purely reactive or uniform spraying strategies.  


\noindent\textbf{Advantages of HAM-PPO Algorithm Over Baseline Models in Fields with Varying Infection Extents:} We evaluate the performance of the HAM-PPO algorithm across fields with varying infection levels and compare its effectiveness to baseline models such as LawnMower+Carpet Spray and LawnMower+HAM-PPO Spray. Specifically, we analyze whether HAM-PPO maintains efficiency as infection levels exceed 60\% and its performance under extreme high-infection scenarios. We use the best policy derived from optimizing for a scenario with low randomness and 20-30\% infection range, where infection initiation occurs at the corners. 
 
\begin{figure}[H]
    \centering
    \includegraphics[width=1.0\linewidth]{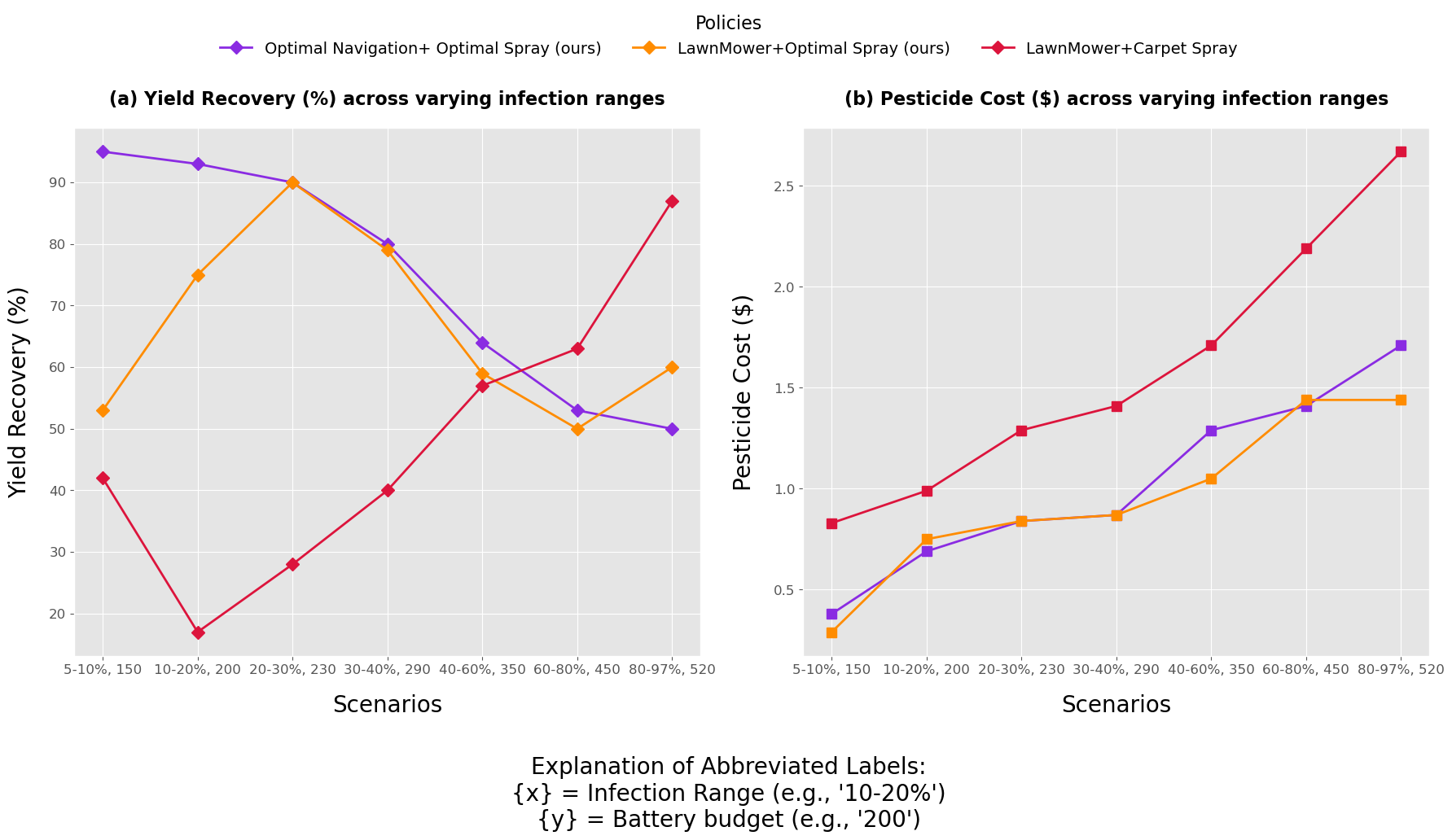}
    \caption{Performance of HAM-PPO and baseline models across varying infection levels. (a) Yield recovery rates across different infection scenarios. (b) Spraying costs incurred by each model.}
    \label{fig:extendedrange}
\end{figure}

As shown in Figure \ref{fig:extendedrange}(a), HAM-PPO demonstrates strong performance in yield recovery rates for infection levels up to 60\%, consistently outperforming LawnMower+Carpet Spray. For example, in fields with 20-30\% infection and a battery budget of 230, HAM-PPO achieves a recovery rate of 0.93, compared to 0.28 for LawnMower+Carpet Spray and 0.9 for LawnMower+HAM-PPO Spray. However, as infection levels rise beyond 60\%, HAM-PPO’s performance decreases. For instance, in the 80-97\% infected field with a battery budget of 520, LawnMower+Carpet Spray achieves a recovery rate of 0.87, outperforming HAM-PPO’s 0.5. 
Importantly, when we extend the evaluation to a very low infection range (5–10\%), a significant performance gap emerges: HAM-PPO maintains very high yield recovery (with values around 95–93\%), whereas LawnMower+HAM-PPO Spray exhibits considerably lower recovery rates, closely aligning with the performance of LawnMower+Carpet Spray. This observation suggests that as the infection level decreases, the benefits of our hierarchical approach become more pronounced.

The Figure \ref{fig:extendedrange}(b) illustrates the cost of spraying for each model, measured in terms of resource usage. HAM-PPO demonstrates its resource efficiency by maintaining consistently lower spraying costs compared to LawnMower+Carpet Spray, particularly in fields with moderate infection levels. For instance, in fields with 40-60\% infection and a battery budget of 350, HAM-PPO incurs a spraying cost of 1.29, compared to 1.71 for LawnMower+Carpet Spray. This highlights HAM-PPO’s ability to balance yield recovery and resource usage effectively. However, in high-infection scenarios, such as 80-97\% infection, where nearly all grids require coverage, LawnMower+Carpet Spray incurs higher costs but achieves a broader recovery due to its full-field coverage strategy.

\begin{figure}[H]
    \centering
    \includegraphics[width=0.9\linewidth]{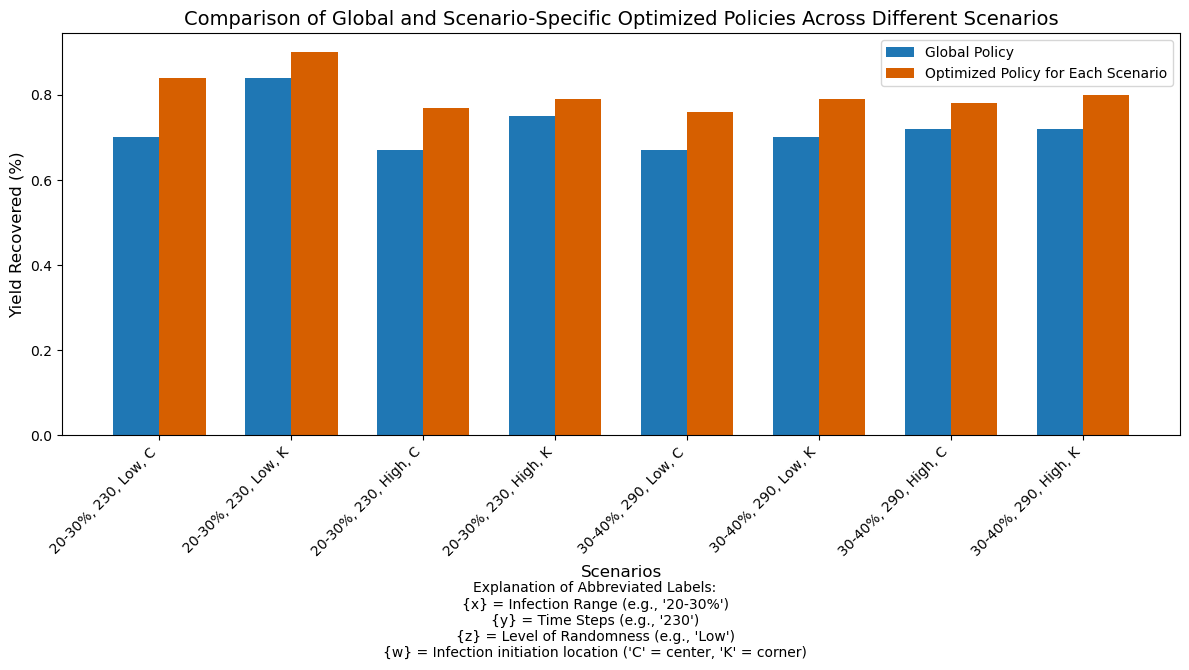}
    \caption{Comparison of global and scenario specific optimized policies across different environmental scenarios.}
    \label{fig:global_policies}
    
\end{figure}

\noindent\textbf{Generalization Capability of the Global Policy:} We compared the global policy, which was trained across all scenarios, with the best policy derived from optimizing for each single specific scenario. While there are slight improvements with optimization, the global policy shows remarkable robustness, with only minimal differences in yield recovery across the tested scenarios. This suggests that the global policy is effective in generalizing across different types of infections and randomness levels, as reflected in Figure \ref{fig:global_policies}.

\begin{figure}[H]
    \centering
    \includegraphics[width=0.9\linewidth]{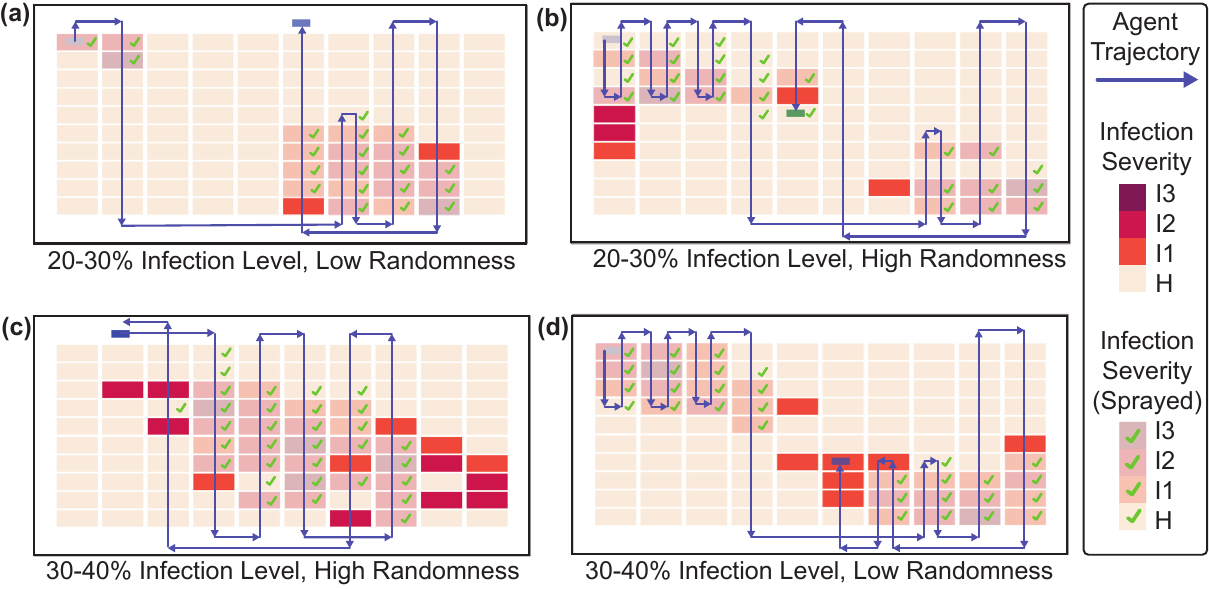}
    \caption{Agent trajectories for four distinct infection scenarios. The subfigures illustrate the HAM-PPO policy's capability to navigate and spray infected plots efficiently.}
    \label{fig:trajectory}
\end{figure}

\noindent\textbf{Trajectory Analysis of HAM-PPO Agent in Varying Infection Scenarios:} The Figure \ref{fig:trajectory} illustrates the agent's trajectories under various distinct infection scenarios, showcasing the HAM-PPO policy's ability to efficiently navigate to infected plots and perform targeted spraying. Across all scenarios, the trajectories demonstrate the policy's adaptability to varying infection patterns, effectively balancing movement and spraying actions. The agent successfully identifies infected regions and sprays them while optimizing its trajectory to minimize unnecessary movements. This highlights the policy's capability to adapt to diverse infection distributions, ensuring efficient resource usage and effective coverage of infected areas.

\section{Conclusions and Future Work}
In this study, we propose a novel approach to optimize navigation and pesticide spraying in precision agriculture using deep reinforcement learning (RL) with a hierarchical action space and conditional action masking. Our method leverages a partially observable environment to simulate the challenges faced by agricultural robots in detecting and spraying infected crops. This approach is particularly useful in environments where there is limited knowledge about the underlying model and where the design of the observation space is crucial to accurately capturing the state of the environment. By incorporating a hierarchical action structure, the agent is able to efficiently scout the field, navigate between grids, and make informed decisions on when and where to spray pesticides, all while managing limited resources such as battery life and pesticide usage.

We demonstrate the effectiveness of our approach in comparison with baseline methods such as LawnMower+Carpet Spray and random policies. The experimental results show that our method, Hierarchical Action Masking Proximal Policy Optimization (HAM-PPO), significantly outperforms these baseline models in terms of yield recovery and resource efficiency. Specifically, HAM-PPO achieves the higher yield recovery percentages under various infection scenarios and demonstrates robustness to environmental randomness. Beyond row-crop agriculture, HAM-PPO has the potential for broader applications in agricultural settings where multi-stage intervention and adaptive resource allocation are critical, such as tree crop orchards~\cite{mahmud2021opportunities} and indoor vertical farming~\cite{gnauer2019towards}.

A key limitation of our approach is its static environment, where the infection status of the field is modeled as a fixed map for each day. This setup does not capture the dynamic nature of infection spread, as it fails to account for the evolution of the infection over time. To address this limitation, we plan to extend our algorithm to operate effectively across spatio-temporal environments with varying dynamics. We aim to incorporate our previously developed AgGym simulator \cite{khosravi2024aggym}, which simulates disease spread throughout the growing season. By integrating this simulator, we can model dynamic infection spread, allowing the agent to adapt to changes in the environment over time. This will enhance the agent's ability to manage disease more effectively by accounting for the evolving status of the field.

In addition to this, we aim to test the RL agent in real-world settings, evaluating its performance on actual agricultural robots. This will involve enriching the observation space with real sensor data and refining the action masking process to further improve the decision-making policy. Furthermore, recent studies have demonstrated the feasibility of automated crop health assessment in soybean using deep learning \cite{naik2017real, rairdin2022deep}. Integrating these automated phenotyping techniques with our RL framework could provide continuously updated insights into crop conditions, enabling even more precise and adaptive pesticide application. Moreover, exploring multi-agent systems holds promise for enabling collaborative decision-making, which can lead to more efficient management of the field. With these advancements, we hope to refine our approach and create a framework that is both practical and effective for real-world precision agriculture applications.

\section*{Acknowledgments}
This work was supported by the COALESCE: COntext Aware LEarning for Sustainable CybEr-Agricultural Systems (CPS Frontier \#1954556), AI Institute for Resilient Agriculture (USDA-NIFA \#2021-647021-35329), Smart Integrated Farm Network for Rural Agricultural Communities (SIRAC) (NSF S \& CC \#1952045).







\bibliographystyle{elsarticle-num}
\bibliography{bib}



\end{document}